\documentclass[11pt]{article}

\usepackage[final]{acl}

\usepackage{times}
\usepackage{latexsym}

\usepackage[T1]{fontenc}

\usepackage[utf8]{inputenc}

\usepackage{microtype}

\usepackage{inconsolata}

\usepackage{graphicx}

\usepackage{multirow}
\usepackage{booktabs}
\usepackage{makecell}
\usepackage[most]{tcolorbox}
\usepackage{float}
\usepackage{xcolor}
\usepackage{amsmath}
\usepackage{url}

\newtcolorbox{promptbox}[1]{
    colback=gray!5!white,
    colframe=gray!75!black,
    leftrule=3pt,
    arc=2pt,
    boxrule=0.5pt,
    fontupper=\small\ttfamily,
    enhanced,
    breakable,
    title={\texttt{#1}},
    left=3pt,
    right=3pt,
    top=2pt,
    bottom=2pt,
}

\title{SWORD: Wikidata-based Distortions Reveal Hidden Cross-Lingual Inconsistencies in LLM Factual Error Rejection}

\author{
  Sanghyeok Park$^1$ \quad
  Minji Kang$^1$ \quad
  Hosung Kwak$^2$ \quad
  \textbf{Jinhyuk Yun}$^{1,}$\thanks{~Corresponding author.} \\
  $^1$Soongsil University, Seoul, Republic of Korea \\
  $^2$KAIST, Daejeon, Republic of Korea \\
  \texttt{\{ps2575, min24, jinhyuk.yun\}@soongsil.ac.kr} \\
  \texttt{hoohoona@kaist.ac.kr}
}

\begin{document}
\maketitle
\begin{abstract}
Modern LLMs demonstrate impressive multilingual performance, yet standard benchmarks primarily reward selecting correct answers rather than evaluating genuine factual understanding. We introduce Systematic Wikidata-based Object-Relation Distortion (SWORD), a benchmark that evaluates whether models consistently reject factual errors across languages. SWORD generates syntactically well-formed but factually incorrect statements in eight widely spoken languages through controlled perturbations of Wikidata triples, ranging from random entity substitutions to semantically plausible property-based selections. Our distortion-based evaluation surfaces two critical insights that remain entirely obscured by conventional benchmarks. First, models counterintuitively achieve higher accuracy on semantically plausible distortions than on nonsensical random substitutions, suggesting reliance on distributional familiarity rather than genuine factual verification. Second, models exhibiting comparable baseline accuracy across languages show substantial performance degradation specifically on (East) Asian languages when presented with distorted statements, with cross-lingual performance gaps reaching up to 28 percentage points (49\% relative reduction) in some models. These findings demonstrate that multilingual factual reasoning involves asymmetric capabilities that aggregate accuracy metrics systematically obscure.
\end{abstract}

\section{Introduction}\label{sec:intro}
Large Language Models (LLMs) have demonstrated strong performance across a wide range of tasks, including multilingual question answering, coding, and reasoning~\cite{meta2024llama31, gemini3flash2025, yang2025qwen3}. In recent years, the multilingual capabilities of LLMs have improved rapidly, demonstrating seemingly stable performance across multiple languages on widely-used benchmarks~\cite{openai2024mmmlu, singh2025global}. However, systematic inequalities persist across languages: models exhibit substantially worse performance and generate more problematic outputs in low-resource languages~\cite{blasi2022systematic, shen2024language}.

Existing multilingual benchmarks are designed around multiple-choice selection or answer extraction~\cite{hendrycks2021measuring, conneau2018xnli, hu2020xtreme, singh2025global}. While effective for automated evaluation and successfully quantifying the multilingual ability of models, high benchmark scores on selecting true statements among the candidates do not guarantee the ability to reject false ones. They also provide limited insight into whether models consistently reject false statements across languages. Most benchmarks simply count correctly answered questions without comparing response patterns between languages~\cite{qi2023cross}, leaving cross-lingual inconsistency (where a model rejects an incorrect statement in one language but accepts it in another) largely invisible.

We propose Systematic Wikidata-based Object-Relation Distortion (SWORD; see Figure~\ref{fig:diagram}), a benchmark designed to evaluate multilingual responses to factually incorrect statements. SWORD systematically perturbs Wikidata triples~\cite{vrandevcic2014wikidata} by altering subject and object components, producing syntactically well-formed but factually incorrect statements. Perturbations range from random entity substitutions to property-based distortions with systematically controlled semantic proximity to the original entities, applied consistently across eight languages.

This study focuses on how LLMs respond to false statements in multilingual settings. Multilingual evaluation is not merely a means of expanding coverage, but a critical condition for testing whether factual judgment remains consistent across languages. We find that language-specific patterns are invisible when evaluation is limited to true statements alone, yet systematic evaluation of false statement rejection reveals such patterns clearly. We also analyze how responses to distorted facts vary across models and examine how cross-lingual performance gaps, nearly invisible in standard benchmarks, emerge dramatically under distortion-based evaluation.

This work makes three contributions. SWORD exposes systematic cross-lingual disparities that emerge specifically under distortion: models exhibiting comparable baseline accuracy show performance gaps of up to 28 percentage points (49\% relative reduction) between Western and East Asian languages when rejecting false statements, a pattern invisible to standard multilingual benchmarks (Sections~\ref{subsec:disparity}~and~\ref{subsec:cluster}). SWORD also reveals counterintuitive evaluation signals: models achieve higher accuracy on semantically plausible distortions (property-based) than on nonsensical ones (random), suggesting reliance on distributional familiarity rather than genuine factual verification (Section~\ref{subsec:performance}). These findings are enabled by a fully automated pipeline that systematically controls distortion difficulty through semantic proximity using Wikidata's structured constraints, allowing scalable benchmark generation without expert validation (Section~\ref{sec:methods}).

\section{Related Works}\label{sec:related}
Multilingual benchmarks have become essential for evaluating LLMs across diverse languages. Early efforts such as GLUE~\cite{wang2018glue} and SuperGLUE~\cite{wang2019superglue} established standardized evaluation frameworks, subsequently extended to multilingual settings through XNLI~\cite{conneau2018xnli} and XTREME~\cite{hu2020xtreme}. More recently, MMLU~\cite{hendrycks2021measuring} and its variants, including MMLU-Pro~\cite{wang2024mmlu}, Multilingual MMLU~\cite{openai2024mmmlu}, and Global MMLU~\cite{singh2025global}, have been introduced. However, these benchmarks evaluate answer selection from candidate pools rather than the absolute judgment of statement veracity, measuring whether models identify the most plausible option among alternatives rather than genuinely verify facts~\cite{mousavi2024dyknow}. Moreover, many remain English-centric, relying on translated data that may not reflect linguistically diverse knowledge~\cite{singh2025global, kassner2021multilingual} or may introduce artifacts that obscure true cross-lingual transfer~\cite{rajaee2024analyzing}.

Recent work has examined cross-lingual consistency more directly. The Ranking-based Consistency (RankC) metric evaluates whether models produce consistent rankings of factual statements across languages~\cite{qi2023cross}, extended to multimodal settings across 15 languages~\cite{wang2025traveling}. Other work proposes translate-then-evaluate strategies for 30 languages~\cite{gupta2025found}, and studies have demonstrated difficulty in maintaining consistent factual predictions across multilingual inputs~\cite{fierro2022factual}. Large-scale work on multilingual hallucination detection, including Poly-FEVER~\cite{zhang2025polyfever} and cross-lingual hallucination rate estimation~\cite{ulislam2025hallucinate}, reveals that smaller models and those supporting more languages show higher hallucination rates. While these studies go beyond aggregate accuracy, they still primarily measure retrieval of correct information rather than consistent rejection of false statements, and lack systematic control over distortion difficulty that would allow distinguishing obviously wrong answers from semantically plausible falsehoods.

Structured knowledge graphs offer a more systematic foundation for factual evaluation. Previous benchmarks extract factual statements from knowledge bases with expert validation~\cite{gupta2021x, augenstein2019multifc}, while others construct large-scale datasets through automated Wikidata sampling~\cite{veseli2023evaluating} or evaluate time-sensitive factual knowledge~\cite{mousavi2024dyknow}. The importance of controlled perturbations has also been recognized: contrast sets test model decision boundaries through minimal label-changing modifications~\cite{gardner2020evaluating}, and recent benchmarks such as FACTS~\cite{cheng2025facts} and FactBench~\cite{bayat2025factbench} emphasize multi-dimensional factuality evaluation. However, existing approaches assess difficulty post-hoc based on model performance rather than controlling semantic proximity a priori, and do not evaluate cross-lingual consistency in false statement rejection. SWORD addresses both gaps through property-based entity substitution with graph-structure-controlled semantic proximity, applied systematically across eight languages.

\section{SWORD Benchmark Construction}\label{sec:methods}

\begin{figure}[!h]
  \includegraphics[width=\columnwidth]{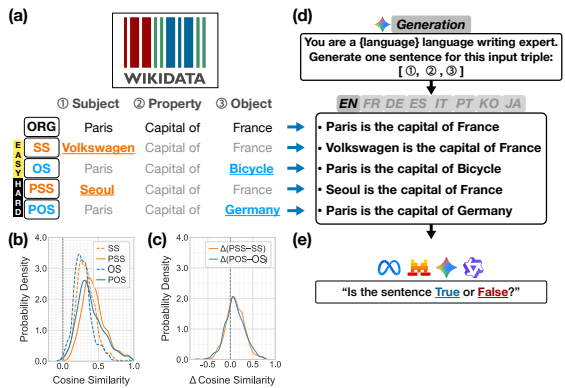}\vspace{-3mm}
  \caption{The SWORD benchmark pipeline generates controlled factual distortions by shuffling Wikidata triples (Subject-Property-Object).} (a) We extract triples from Wikidata (true statements) and apply four shuffling strategies to generate false statements (SS: Subject Shuffle, PSS: Property-based Subject Shuffle, OS: Object Shuffle, POS: Property-based Object Shuffle; see Methods for details). (b) The probability density plot shows cosine similarities between original and shuffled entities for each strategy by node2vec embedding. (c) Difference in cosine similarity between the random--original pair and property-based--original pair. Property-based shuffles tend to produce entities with higher similarity to the original, demonstrating that semantic proximity is directly linked to task difficulty and enabling systematic creation of easy-to-hard evaluation tasks. (d) Both original and shuffled triples are converted into natural language statements in 8 languages using Gemini-2.5-Flash. (e) LLMs are then asked to judge whether each statement is factually true or false.
  \label{fig:diagram}\vspace{-5mm}
\end{figure}

SWORD evaluates cross-lingual factual accuracy of LLMs by measuring their ability to both accept true statements and reject systematically distorted false ones. Our benchmark generates two types of triples from Wikidata $S$-$P$-$O$ triples ($S$: Subject, $P$: Property, $O$: Object): (i) original true relations and (ii) systematically distorted false relations (Figure~\ref{fig:diagram}(a--c)). Each triple is translated into a natural language sentence in each of eight widely spoken languages (EN, FR, DE, ES, IT, PT, KO, JA) using Gemini-2.5-Flash (Figure~\ref{fig:diagram}(d))~\cite{gemini3flash2025}, and models are asked to classify each statement as true or false (Figure~\ref{fig:diagram}(e)). SWORD is applicable to both API-based and open-source models, and the benchmark size can be scaled according to available computational resources. The benchmark and code are publicly available at \url{https://anonymous.4open.science/r/SWORD-EMNLP/}

\subsection{Data Collection and Systematic Distortion}\label{subsec:distorsion}

We use a Wikidata dump containing data up to September 11, 2024. Our extraction starts from each Wikidata entity as a subject and linked objects through properties to construct $S$-$P$-$O$ triples. The dump contains $112,431,677$ entities (items) and $12,120$ properties, composing $788,124,878$ triples in total. We focus on obtaining original triples (denoted \texttt{ORG}) that can be translated into standalone factual sentences. To ensure fair cross-lingual comparison, we retain only triples whose subject, property, and object all have natural-language labels in all eight target languages. This constraint ensures that selected entities are likely to appear in training corpora across all languages, reducing potential bias from language-specific entity coverage. After this filtering, our final dataset contains $2,119,631$ triples.

We then sample 500 triples to test our benchmark. For each \texttt{ORG} triple $(S, P, O)$, we generate four types of distorted triples by keeping the property $P$ fixed and substituting either the subject $S$ or the object $O$ (Figure~\ref{fig:diagram}(a)). This produces false statements that remain syntactically well-formed while varying in semantic plausibility.

\vspace{-3mm}
\begin{table}[h]
\centering
\caption{Four distortion types from each \texttt{ORG} triple.}
\label{tab:distortion_types}
\vspace{-2mm}
\small
\setlength{\tabcolsep}{4pt}
\begin{tabular}{ll}
\toprule
\textbf{Type} & \textbf{Description} \\
\midrule
\texttt{ORG} & Original factual triple from Wikidata \\
\texttt{SS} & Replace $S$ with a random entity $S'$ \\
\texttt{OS} & Replace $O$ with a random entity $O'$ \\
\texttt{PSS} & Replace $S$ with $S'$ sharing property $P$ with same $O$ \\
\texttt{POS} & Replace $O$ with $O'$ sharing property $P$ with same $S$ \\
\bottomrule
\end{tabular}
\vspace{-3mm}
\end{table}

To ensure that each distorted triple represents a false relation, we verify whether the resulting triple exists in Wikidata. Since Wikidata contains many-to-many relations, random shuffling may accidentally produce existing triples; in such cases, we resample until obtaining a non-existent triple, guaranteeing that all shuffled triples are labeled \texttt{False}.

To verify that property-based shuffling produces more semantically plausible distortions than random shuffling, we apply node2vec~\cite{perozzi2014deepwalk, grover2016node2vec} to an undirected graph constructed from the $2,119,631$ filtered Wikidata triples, using a walk length of $10$ steps, $10$ walkers per node, and skip-gram with negative sampling. We then measure cosine similarity between original and shuffled entities to confirm that property-based strategies yield semantically closer substitutions (Figure~\ref{fig:diagram}(b,c)).

Figures~\ref{fig:diagram}(b,c) confirm that property-based strategies (PSS and POS) consistently produce semantically closer substitutions than random strategies (SS and OS), validating that property-based distortions create more plausible false statements while random shuffling produces obviously nonsensical ones. The entire pipeline, from triple extraction and distortion generation to multilingual statement creation, is fully automated, requiring no expert validation or manual curation.

\subsection{Multilingual Statement Generation}\label{subsec:generation}
We convert each \texttt{ORG} triple and its four distortions into a single declarative sentence in each of eight languages using Gemini-2.5-Flash~\cite{gemini3flash2025}. For reproducibility, we use the default settings with a fixed prompt template (see Appendix~\ref{sec:prompts}).

We use English instructions for all target languages, as multilingual LLMs internally perform key reasoning in representation spaces closest to English regardless of input and output languages~\cite{schut2025multilingual}. For each target language, we instantiate the same template with the corresponding Wikidata labels, explicitly requesting exactly one sentence to keep outputs consistent. This yields $2,500$ statements per language and $20,000$ statements across eight languages.

\subsection{True/False Classification Evaluation}\label{subsec:eval}
Given each generated statement in a target language, we query a model and ask it to judge whether the statement is \textbf{True} or \textbf{False}, using again a single fixed evaluation prompt in English for all models~\cite{schut2025multilingual} to ensure performance differences reflect inherent capabilities rather than prompt variations (see Appendix~\ref{sec:prompts}). We parse the model output by extracting the label inside the \texttt{<answer>} tags (case-insensitive); if the tags are missing or the output does not match one of the allowed labels, we map the response to \texttt{Unsure}. For scoring, \texttt{ORG} statements serve as ground-truth \textbf{True} labels, while all distorted statements (\texttt{SS}, \texttt{OS}, \texttt{PSS}, \texttt{POS}) serve as ground-truth \textbf{False} labels. We treat \texttt{Unsure} responses as incorrect, as they indicate failure to provide a definitive judgment as instructed. This choice does not affect our core findings: LLaMA and Mistral show virtually zero \texttt{Unsure} responses (5-run avg. 0.47 and 0.03 out of 500, respectively), meaning their cross-lingual gaps are driven entirely by incorrect responses, while Qwen3 models (non-think mode) show consistently distributed \texttt{Unsure} rates across languages, introducing no systematic cross-lingual bias (see Appendix Figure~\ref{fig:heatmap_spectrum} and Tables~\ref{tab:accuracy_by_lang_with_unsure} and~\ref{tab:accuracy_by_lang_wo_unsure}). We compute accuracy separately for each statement type and language, yielding $20,000$ statements in total.

\section{Cross-Lingual Evaluation Reveals Hidden Performance Disparities}\label{sec:results}
\begin{figure*}[t]
    \centering
    \includegraphics[width=0.9\textwidth]{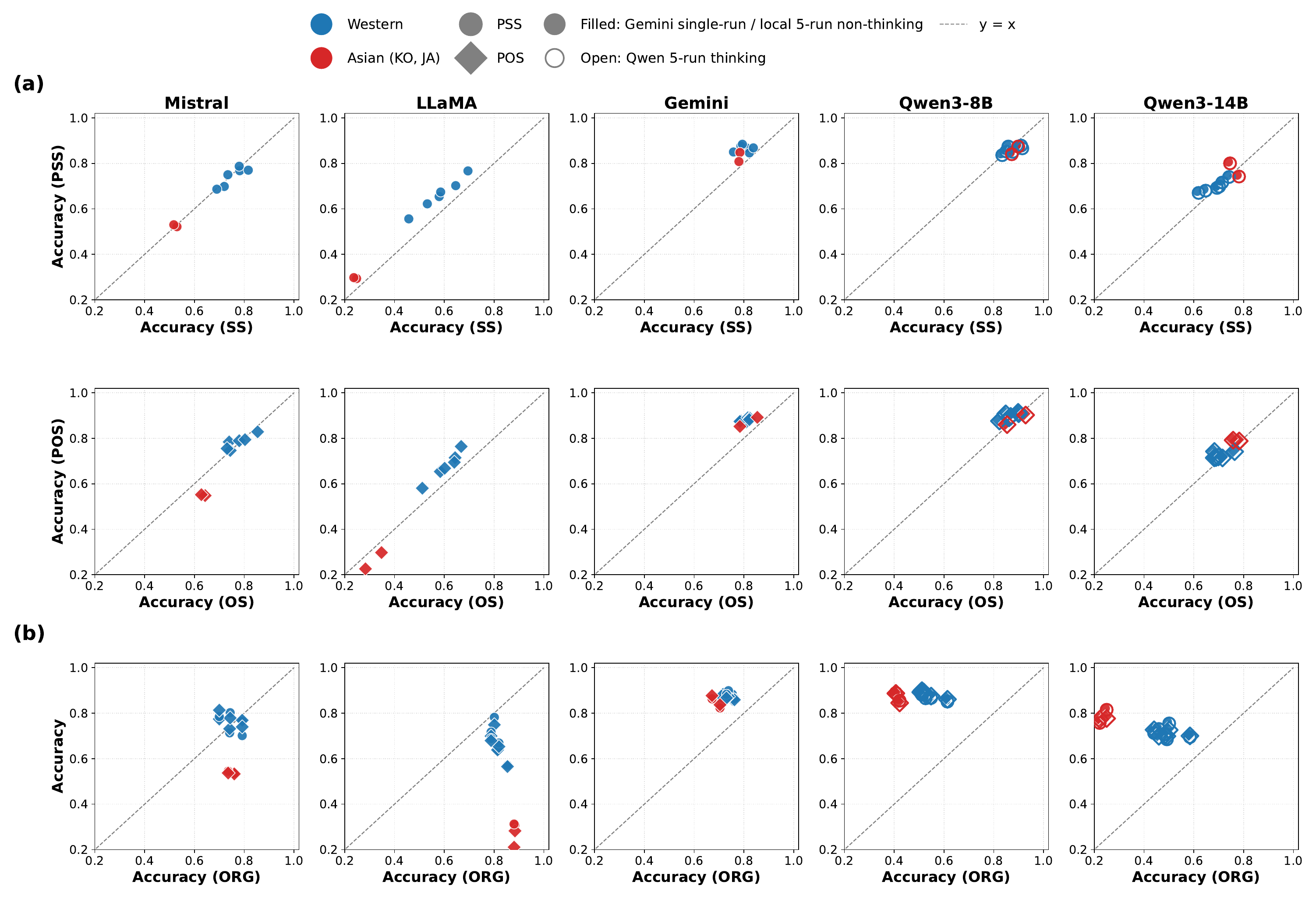}\vspace{-3mm}
        \centering   
        \caption{Comparison of factual accuracy across models and languages under different distortion levels. Each column corresponds to a model, and dot colors indicate language groups (Western: blue; Asian (Korean and Japanese): red). (a) PSS vs. SS (first row) and POS vs. OS (second row) accuracy comparison. (b) Property-based shuffling strategies (PSS and POS) versus original statements (ORG). The diagonal line ($y = x$) represents equal performance across compared conditions. Counterintuitively, random shuffles show lower accuracy than property-based shuffles across most models, suggesting that semantically plausible distortions are easier to reject than nonsensical ones. For LLaMA 3.1 and Mistral Nemo, Asian languages show more severe performance degradation on shuffled statements than on original statements, a disparity not observed in Gemini or Qwen3.}
        \label{fig:accuracy}\vspace{-5mm}
\end{figure*}

We evaluate five models (Mistral-Nemo-Instruct-2407, LLaMA-3.1-Instruct-8B, Gemini-2.5-Flash, Qwen3-8B, Qwen3-14B) on 500 randomly sampled \texttt{ORG} triples and their corresponding four distortions. We use the official Gemini API for Gemini-2.5-Flash (single run due to API cost constraints) and vLLM for all other models; local models are evaluated over 5 random seeds and report mean accuracy. Given the negligible variance observed across local models (std. $\leq$ 0.002), single-run Gemini results are expected to be reliable. In this section, Qwen3 results are reported in non-think mode, yet think mode yields consistent cross-lingual patterns (Figure~\ref{fig:accuracy}). Note that our main evaluation targets are small local models. Gemini is included as a baseline since it generated the statements and thus benefits from evaluator self-preference~\citep{panickssery2024llm}. We organize our analysis into three parts: counterintuitive performance patterns on distorted statements (Section~\ref{subsec:performance}), cross-lingual performance disparities (Section~\ref{subsec:disparity}), and multilingual consistency analysis (Section~\ref{subsec:cluster}).

\begin{figure*}[h]
    \centering
    \includegraphics[width=0.9\textwidth, trim={3.5cm 2.5cm 3.5cm 1.5cm}, clip]{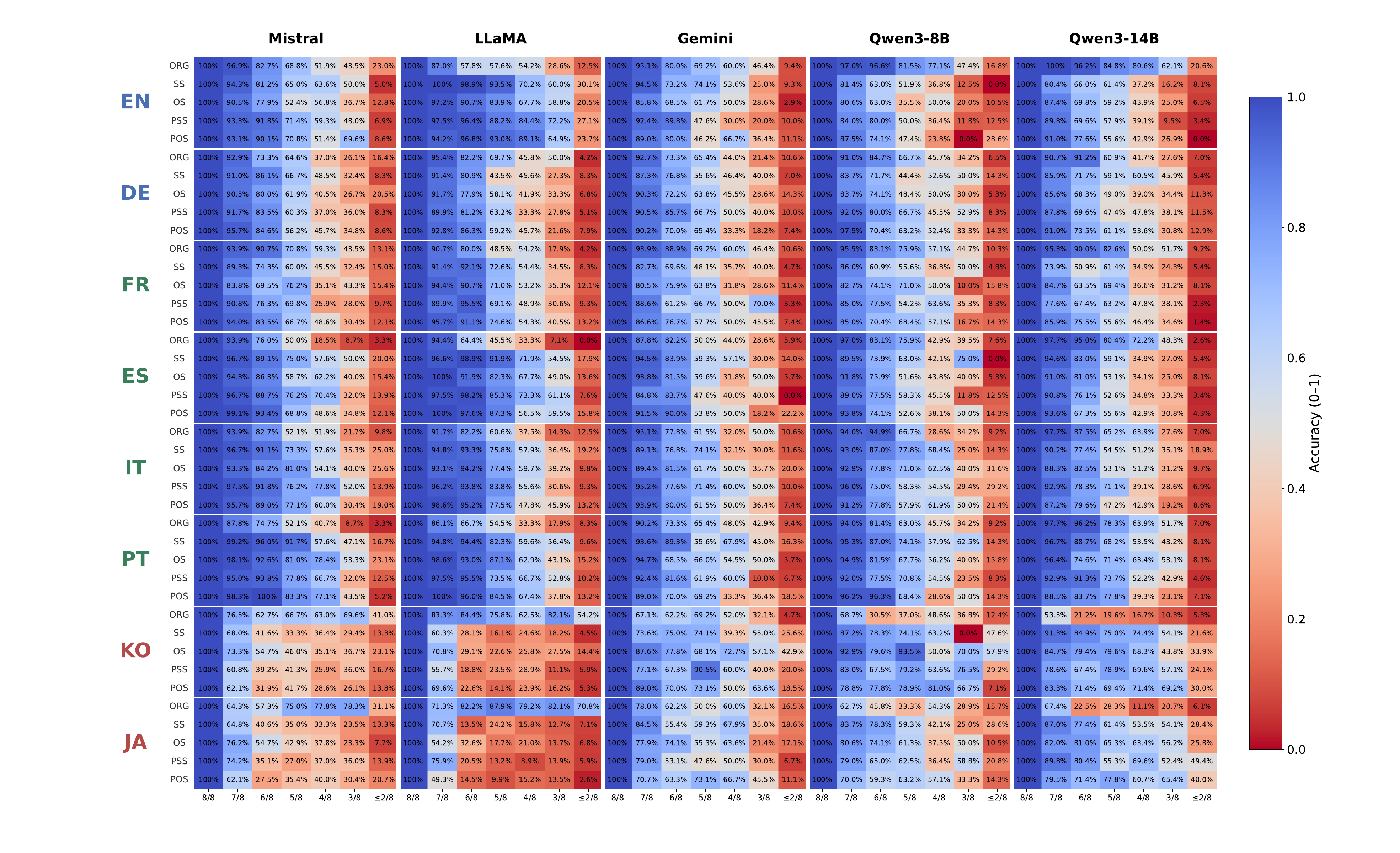} \vspace{-3mm}
    \caption{Accuracy distribution across models and languages grouped by problem difficulty, measured by the number of languages (out of 8) in which the \texttt{ORG} statement is correctly classified. The leftmost bin ($8/8$) contains the easiest statements (correct in all languages), while difficulty increases rightward ($\leq 2/8$: correct in 2 or fewer languages). Rows are grouped by language family: Germanic languages (EN, DE; though English has substantial Romance influence) at the top, Romance languages (FR, ES, IT, PT) in the middle, and Asian languages (KO, JA) at the bottom. Shuffled performance degrades more severely for Asian languages as difficulty increases, particularly in LLaMA 3.1 and Mistral Nemo, while Gemini maintains consistent cross-lingual performance across all difficulty levels, and Qwen3's apparent consistency is better attributed to its response bias toward False than to genuine cross-lingual stability. This bias may contribute to Qwen3’s ORG performance degradation in Asian languages.}
    \label{fig:heatmap}
    \vspace{-5mm}
\end{figure*}

\begin{table}[t]
\centering
\caption{Accuracy across models, shuffle types, and languages. All local models report five-run mean accuracy; Gemini-2.5-Flash reports single-run results. Qwen3 results are in non-think mode. Unsure responses are counted as incorrect.}
\label{tab:accuracy_by_lang}
\vspace{-3mm}
\scriptsize
\setlength{\tabcolsep}{2.0pt}
\renewcommand{\arraystretch}{0.92}
\resizebox{\columnwidth}{!}{%
\begin{tabular}{llcccccccc}
\toprule
\textbf{Model} & \textbf{Type} &
\textbf{EN} & \textbf{DE} & \textbf{FR} & \textbf{ES} &
\textbf{IT} & \textbf{PT} & \textbf{KO} & \textbf{JA} \\
\midrule
\multirow{5}{*}{\makecell{Mistral\\-Nemo\\-Instruct\\-2407}} 
 & ORG & \textbf{0.792} & 0.742 & \textbf{0.792} & 0.700 & 0.744 & 0.700 & 0.760 & 0.736 \\
 & SS & 0.734 & 0.720 & 0.690 & 0.782 & 0.780 & \textbf{0.816} & 0.530 & 0.517 \\
 & OS & 0.740 & 0.744 & 0.731 & 0.780 & 0.803 & \textbf{0.854} & 0.643 & 0.628 \\
 & PSS & 0.750 & 0.698 & 0.686 & 0.768 & \textbf{0.787} & 0.770 & 0.522 & 0.530 \\
 & POS & 0.784 & 0.746 & 0.755 & 0.789 & 0.794 & \textbf{0.828} & 0.548 & 0.552 \\
\midrule

\multirow{5}{*}{\makecell{LLaMA\\-3.1\\-Instruct\\-8B}} 
 & ORG & 0.801 & 0.854 & 0.814 & 0.788 & 0.820 & 0.788 & \textbf{0.884} & 0.880 \\
 & SS & \textbf{0.695} & 0.458 & 0.532 & 0.646 & 0.580 & 0.586 & 0.248 & 0.237 \\
 & OS & \textbf{0.668} & 0.512 & 0.584 & 0.644 & 0.602 & 0.640 & 0.348 & 0.284 \\
 & PSS & \textbf{0.767} & 0.556 & 0.622 & 0.702 & 0.654 & 0.674 & 0.294 & 0.298 \\
 & POS & \textbf{0.764} & 0.580 & 0.654 & 0.714 & 0.668 & 0.694 & 0.298 & 0.226 \\
\midrule

\multirow{5}{*}{\makecell{Gemini\\-2.5\\-Flash}}
 & ORG & 0.754 & 0.722 & \textbf{0.762} & 0.710 & 0.738 & 0.730 & 0.672 & 0.704 \\
 & SS & 0.810 & 0.788 & 0.758 & 0.822 & 0.794 & \textbf{0.838} & 0.784 & 0.780 \\
 & OS & 0.788 & 0.810 & 0.784 & 0.816 & 0.824 & 0.822 & \textbf{0.854} & 0.784 \\
 & PSS & 0.868 & 0.876 & 0.850 & 0.846 & \textbf{0.884} & 0.868 & 0.848 & 0.808 \\
 & POS & 0.876 & 0.872 & 0.874 & 0.890 & 0.888 & 0.882 & \textbf{0.892} & 0.852 \\
\midrule

\multirow{5}{*}{\makecell{Qwen3\\-8B\\Non-think}}
 & ORG & \textbf{0.608} & 0.500 & 0.543 & 0.521 & 0.513 & 0.505 & 0.401 & 0.416 \\
 & SS & 0.828 & 0.853 & 0.844 & 0.871 & 0.904 & \textbf{0.908} & 0.892 & 0.866 \\
 & OS & 0.817 & 0.842 & 0.854 & 0.862 & 0.894 & 0.892 & \textbf{0.921} & 0.847 \\
 & PSS & 0.842 & 0.880 & 0.858 & 0.856 & \textbf{0.884} & 0.872 & 0.880 & 0.846 \\
 & POS & 0.882 & 0.915 & 0.896 & 0.902 & 0.914 & \textbf{0.920} & 0.908 & 0.866 \\
\midrule

\multirow{5}{*}{\makecell{Qwen3\\-14B\\Non-think}}
 & ORG & \textbf{0.580} & 0.434 & 0.486 & 0.478 & 0.454 & 0.494 & 0.216 & 0.245 \\
 & SS & 0.641 & 0.695 & 0.614 & 0.688 & 0.708 & 0.735 & \textbf{0.774} & 0.739 \\
 & OS & 0.683 & 0.676 & 0.677 & 0.694 & 0.710 & 0.758 & \textbf{0.776} & 0.752 \\
 & PSS & 0.686 & 0.704 & 0.677 & 0.698 & 0.722 & 0.746 & 0.748 & \textbf{0.806} \\
 & POS & 0.722 & 0.747 & 0.720 & 0.724 & 0.720 & 0.746 & 0.794 & \textbf{0.799} \\
\bottomrule
\end{tabular}
}
\vspace{-6mm}
\end{table}

\begin{figure*}[h]
    \centering
    \includegraphics[width=0.65\textwidth]{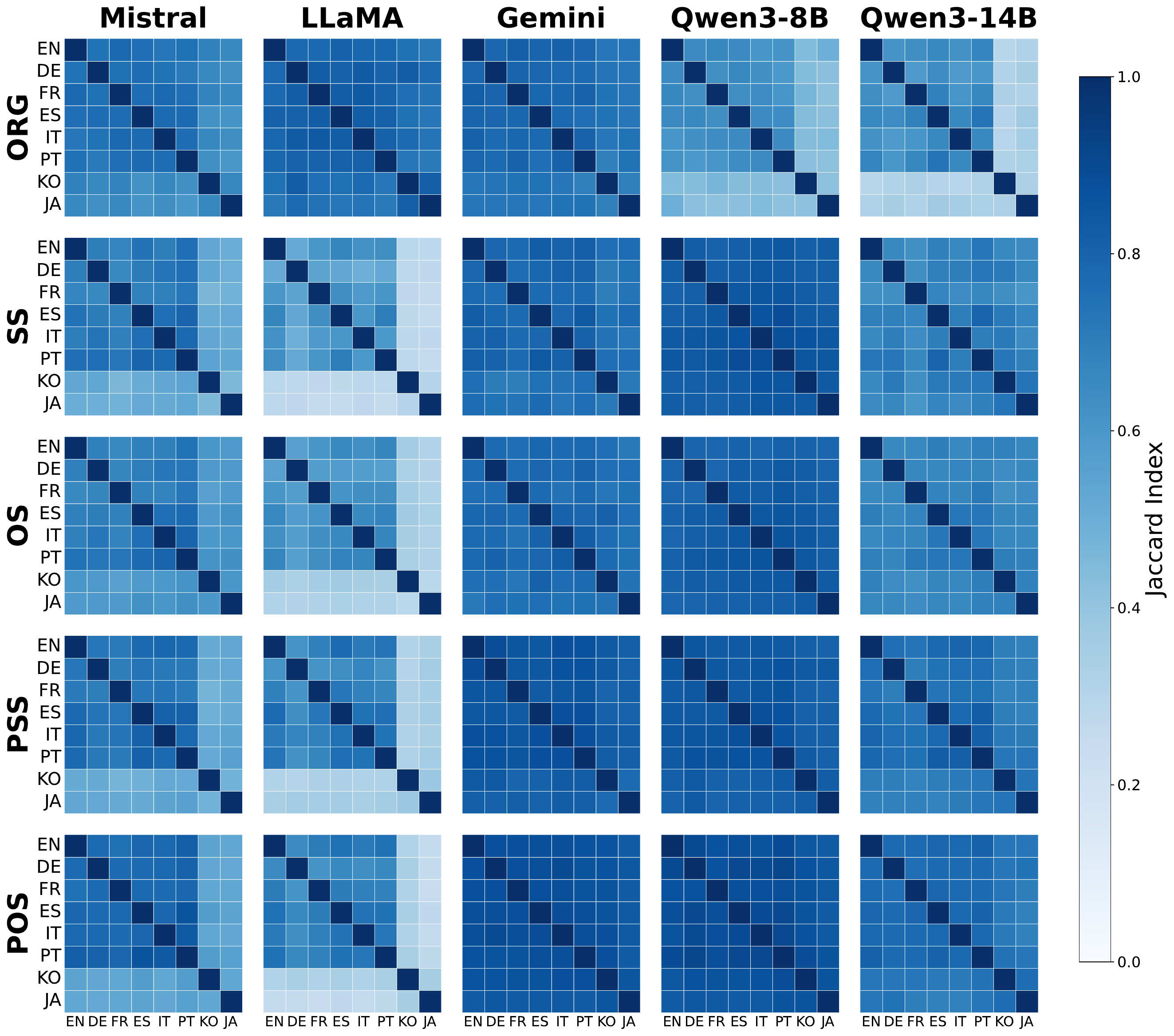}
    \caption{Language-language Jaccard similarity based on correctly answered statements, where each cell shows the proportion of statements correctly answered in both languages relative to those correctly answered in at least one language. Higher values (darker blue) indicate greater overlap. For \texttt{ORG} statements, most models show weak language clustering. Qwen3 models are the exception, showing visible Western language clustering due to lower accuracy on Asian languages. Under distortions, LLaMA 3.1 and Mistral Nemo develop strong Western/Asian language family blocks, while Qwen3 models show the opposite: their Western clustering disappears under distortion, though this reflects Qwen3's response bias toward False rather than genuine cross-lingual stability. Gemini 2.5 Flash maintains weak clustering across all conditions, suggesting uniformly stable cross-lingual factual reasoning.}
    \label{fig:jaccard_lang}
    \vspace{-5.5mm}
\end{figure*}

\begin{figure*}[h]
    \centering
    \includegraphics[width=0.85\textwidth]{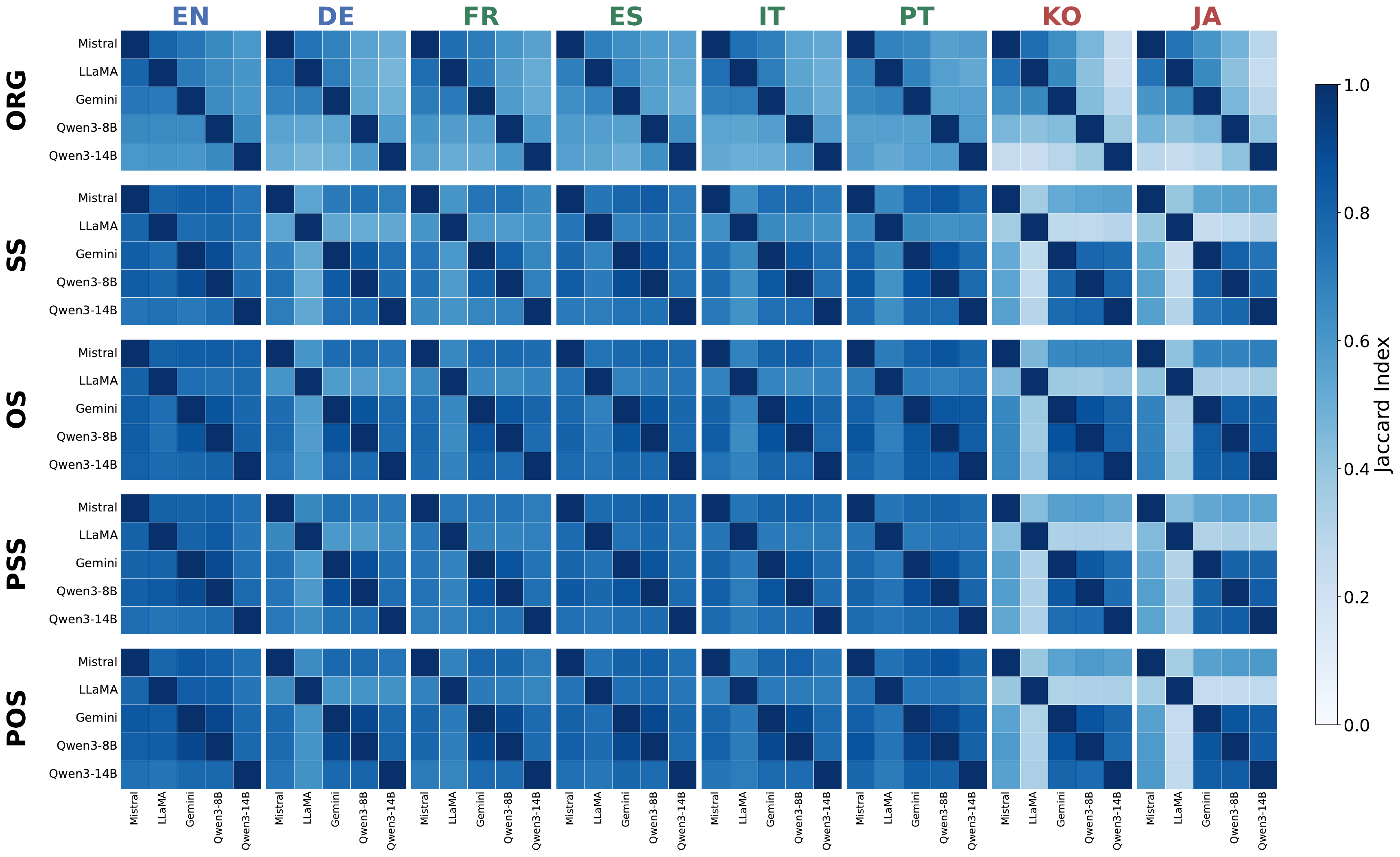}
    \vspace{-3mm}
    \caption{Model-model Jaccard similarity based on correctly answered statements across languages (similarity metric as in Figure~\ref{fig:jaccard_lang}). Each column block corresponds to a language, and each row shows a different condition. For \texttt{ORG} statements, models show relatively uniform similarity across Western languages, but Qwen3-14B exhibits lower agreement in Asian languages (KO, JA) due to its lower accuracy on \texttt{ORG} statements. Under distortions, LLaMA 3.1 shows substantially lower similarity with all other models across all languages, with this pattern most pronounced in Korean and Japanese, revealing that factual distortions amplify model-specific reasoning patterns in Asian languages that remain invisible in \texttt{ORG} statements.}
    \label{fig:jaccard_model} 
    \vspace{-5mm}
\end{figure*}

\subsection{Plausible Distortions Are Easier to Reject than Nonsensical Ones}\label{subsec:performance}
On \texttt{ORG} statements, model performance follows patterns consistent with existing multilingual benchmarks (Table~\ref{tab:accuracy_by_lang}; cf. MMMLU in Table~\ref{tab:mmlu_multilingual}), with accuracy broadly following English $>$ European $>$ Asian languages. However, performance on shuffled statements reveals counterintuitive patterns.

Figure~\ref{fig:accuracy}(a) and Table~\ref{tab:accuracy_by_lang} compare model performance on random versus property-based distortions. Across 80 combinations (5 models $\times$ 8 languages $\times$ 2 distortion pairs), 59 cases (73.75\%) show equal or higher accuracy on property-based shuffles than on random shuffles. This pattern contrasts the expectation that semantically plausible false statements (property-based) would be harder to reject than nonsensical ones (random). As demonstrated in Figure~\ref{fig:diagram}(b,c), property-based shuffles produce entities with higher semantic similarity to the originals, yet models generally find them easier to reject. This suggests that nonsensical random substitutions fall outside the models' training distribution, leading to unstable behavior rather than confident rejection.

This pattern varies by model. LLaMA 3.1 and Gemini consistently show higher accuracy on property-based distortions (14/16 and 16/16 cases), while Mistral Nemo shows no clear preference (8/16 cases each direction). Qwen3 models perform better on property-based subject shuffles but show mixed patterns for object shuffles, possibly reflecting differences between Wikidata's knowledge structure and each model's learned representations.

\subsection{Distortion Exposes Severe Cross-Lingual Disparities in Asian Languages}\label{subsec:disparity}

For LLaMA 3.1 and Mistral Nemo, Asian languages (KO, JA) fall substantially below the diagonal line in Figure~\ref{fig:accuracy}, indicating sharp performance degradation on shuffled statements that far exceeds what existing multilingual benchmarks suggest. For LLaMA 3.1, MMMLU shows only a $\sim$5 percentage point gap between German (55.63) and Korean (50.78) (Table~\ref{tab:mmlu_multilingual}), and on \texttt{ORG} statements Korean even slightly outperforms German (0.884 vs. 0.854). However, under property-based shuffles the gap widens dramatically: German achieves 0.556 (\texttt{PSS}) and 0.580 (\texttt{POS}), while Korean drops to 0.294 and 0.298 respectively, a gap of 26--28 percentage points (47--49\% relative reduction). When averaged across languages, the five European languages achieve 0.641 on \texttt{PSS} compared to 0.296 for Asian languages, whereas MMMLU scores show only 57.8 versus 51.4~\cite{multilingual_mmlu_leaderboard}.

Mistral Nemo shows a similar pattern. MMMLU suggests only a 6\% gap between European languages (62.84\%) and Japanese (59\%), yet under property-based shuffles European languages achieve 0.743 (\texttt{PSS}) and 0.782 (\texttt{POS}) compared to Japanese's 0.530 and 0.552, gaps of 21--23 percentage points (29--30\% relative reduction). Critically, this disparity does not appear in Gemini or Qwen3 models, confirming it is model-specific rather than inherent to the task or language family. 

For statement-level analyses (Figures~\ref{fig:heatmap}--\ref{fig:jaccard_model}), we use a single representative run (seed 5555) under non-think mode for all local models. Think and non-think modes yield consistent cross-lingual patterns (Figure~\ref{fig:accuracy}). Variance across seeds is negligible (std. $\leq$ 0.002, Table~\ref{tab:accuracy_by_lang}), so results generalize across seeds and reasoning modes. The unbinned visualization in Figure~\ref{fig:heatmap_spectrum} confirms these patterns across the full spectrum of difficulty, independently of bin boundaries.

Figure~\ref{fig:heatmap} further examines this disparity by grouping statements by difficulty. Asian language degradation intensifies for harder problems in LLaMA and Mistral, while Gemini maintains consistent cross-lingual performance across all difficulty levels, and Qwen3's apparent consistency is better attributed to its response bias toward False than to genuine cross-lingual stability. These observations demonstrate that SWORD's distortion-based evaluation reveals cross-lingual disparities invisible to standard benchmarks like MMMLU, highlighting the importance of false statement rejection in multilingual evaluation.

\subsection{Distortions Reveal Language Family Clustering in Model Responses}\label{subsec:cluster}

We investigate whether models show consistent response patterns across languages using Jaccard similarity on correctly answered statements, comparing agreement across languages within each model (Figure~\ref{fig:jaccard_lang}) and agreement between models within each language (Figure~\ref{fig:jaccard_model}).

For \texttt{ORG} statements, most models show relatively weak language clustering. Qwen3 models are the exception, showing visible Western language clustering due to notably lower accuracy on Asian languages in \texttt{ORG} statements.

Under distortions, the patterns diverge sharply. LLaMA 3.1 and Mistral Nemo, which showed no clustering on \texttt{ORG} statements, develop strong Western/Asian language family blocks under distortion, reflecting their severe performance degradation on Asian languages specifically when rejecting false statements. Conversely, Qwen3 models show the opposite: the Western clustering present in ORG statements largely disappears under distortion. However, given Qwen3's strong bias toward rejecting statements (reflected in notably low ORG accuracy in Asian languages), this pattern is better interpreted as a consequence of response bias rather than genuine cross-lingual stability. Gemini 2.5 Flash maintains consistently weak clustering across all conditions, suggesting uniformly stable cross-lingual factual reasoning.

Comparing agreement between models within each language (Figure~\ref{fig:jaccard_model}), English exhibits consistently high cross-model agreement across all conditions. Under distortions, LLaMA 3.1 exhibits particularly low cross-model agreement in all non-English languages, most significantly in Korean and Japanese, revealing that factual distortions amplify model-specific reasoning patterns that remain invisible in \texttt{ORG} statements.

These clustering patterns are consistent across reasoning modes; both Qwen3-8B and Qwen3-14B show the same language family structure in think mode (Appendix Figures~\ref{fig:heatmap_think}--\ref{fig:jaccard_model_think}).

\section{Discussion}

SWORD demonstrates that standard evaluation conflates two distinct capabilities: accepting true statements and rejecting false ones. While recent benchmarks increasingly adopt harder problems to maintain discriminative power~\cite{chollet2025arc, rein2024gpqa, phan2025humanity}, SWORD shows that strategic manipulation of simple factual statements can reveal model-specific patterns invisible in aggregate accuracy scores. This suggests that benchmark design should prioritize diagnostic power over raw difficulty.

Our findings carry implications for understanding multilingual factual reasoning in LLMs. The counterintuitive advantage of plausible over nonsensical distortions suggests models rely on distributional familiarity with entity co-occurrences rather than genuine factual verification. The dramatic cross-lingual performance gaps that emerge under distortion, despite comparable baseline accuracy, indicate that multilingual training does not uniformly transfer the ability to reject false statements across language families. That these gaps appear in LLaMA and Mistral but not in Gemini or Qwen3 suggests model-specific factors, particularly training data composition and multilingual alignment strategies, play a critical role. Notably, the similar degradation patterns in Korean and Japanese argue against tokenizer or script complexity as the primary driver; both languages show comparable drops despite different writing systems (Hangul vs. mixed kana/kanji). This interpretation is further supported by preliminary results on Arabic, where LLaMA shows severe degradation consistent with KO/JA patterns while Mistral does not (Table~\ref{tab:arabic_only_results}), 
suggesting that training data distribution rather than script type underlies the observed disparities. Qwen3 models exhibit notably low \texttt{ORG} accuracy in Korean and Japanese (e.g., Qwen3-14B at 0.216 and 0.245), suggesting  a conservative response bias toward rejecting statements; their high distortion-rejection accuracy in these languages should therefore be  interpreted alongside this baseline asymmetry rather than as evidence of superior factual verification.

While factual tasks in production systems are increasingly handled through retrieval-augmented generation (RAG), evaluating parametric knowledge remains independently valuable. Small-scale local models (8B--14B), widely deployed in resource-constrained and security-sensitive environments, often operate without RAG infrastructure. Moreover, even in RAG-based systems, parametric knowledge underlies factual consistency assessment of retrieved content. SWORD's distortion-based evaluation thus complements RAG-focused benchmarks by isolating cross-lingual variance in parametric knowledge.

These findings have practical implications for LLM evaluation and deployment. Aggregate accuracy on true statements provides insufficient characterization of multilingual reliability, as real-world use requires models to verify both true and false claims. Incorporating false-statement rejection, difficulty-controlled distortions, and cross-lingual consistency measures can enrich evaluation frameworks. SWORD provides a scalable, fully automated means to surface these failure modes prior to deployment.

\clearpage

\section{Limitations}

Our findings have several limitations. First, our main results cover eight languages, and whether the observed cross-lingual disparities generalize to lower-resource languages remains an open question. Preliminary results on Arabic (Appendix Table~\ref{tab:arabic_only_results}) suggest similar patterns extend beyond East Asian languages, though a more systematic analysis across a broader range of languages is needed. Additionally, LLaMA 3.1 does not officially support Korean or Japanese, and Mistral Nemo does not report Korean performance in its model card; the observed degradation on these languages may therefore partly reflect limited language support. However, two observations argue against limited language support as the primary explanation. First, these models are widely deployed in practice regardless of official support boundaries, making their degradation patterns practically relevant. Second, and more critically, Mistral Nemo claims Korean and Japanese support yet exhibits a similar degradation pattern as LLaMA 3.1, suggesting that the observed disparities reflect model-specific training factors rather than the absence of language support. If anything, the lack of language-specific fine-tuning in unsupported languages may make underlying cross-lingual disparities more visible, as post-training adjustments in supported languages could otherwise mask such patterns. Additionally, while we use English evaluation prompts to isolate factual reasoning from multilingual instruction-following capabilities, native-language prompting could yield different performance profiles.

Second, multilingual statement generation relies on Gemini-2.5-Flash, 
which likely contributes to Gemini's strong performance through self-preference effects \citep{panickssery2024llm}; cross-model comparisons involving Gemini should therefore focus on relative patterns across languages rather than absolute accuracy. While Gemini-2.5-Flash was selected for statement generation after comparison with several open-source models, linguistic verification of generation quality was limited to languages accessible to the authors (Korean and English at a high level of proficiency, French and Japanese at a conversational level), and generation quality in the remaining languages could not be directly assessed. Furthermore, while we use knowledge graph embeddings to control semantic proximity, semantic plausibility is multifaceted, and the observed counterintuitive pattern may partly reflect imperfect similarity control.

While the observed effect sizes are large and consistent across models, the generalizability of specific numerical findings to other triple sets or model families warrants caution. Notably, the key findings replicate across five models and all eight languages, and the negligible variance across random seeds (std. $\leq$ 0.002) suggests that 500 triples provide a stable evaluation signal for the effect sizes reported here.

Fourth, we standardized our base triples using a static Wikidata dump from September 2024 to maintain a controlled evaluation environment. Although evaluations involving time-sensitive facts may exhibit variant behavior on more recent model iterations~\cite{mousavi2024dyknow}, this fixed temporal baseline ensures the replicability of our structural distortion analysis.

Finally, while Wikidata is widely acknowledged as the gold standard for large-scale structured knowledge base evaluation, its inherent structural density variations across regions might subtly interact with our findings. Disentangling model-specific factual decay from systemic data imbalances remains an important direction for future work. Both the evaluation format and triple structure also reflect inherent properties of Wikidata. Statements are encyclopedic and entity-centric, and ground truth is binary by construction. Whether these findings generalize to other types of factual knowledge, such as procedural or contextually dependent facts, remains an open question.

Despite these limitations, our findings reveal previously invisible challenges in multilingual AI evaluation that extend beyond accuracy metrics. As AI systems increasingly shape global information access and decision-making, ensuring their reliability across languages becomes not merely a technical problem but a matter of global equity. By separating the ability to accept true statements from the ability to reject false ones, and by exposing how these capabilities diverge across languages, we hope SWORD contributes toward evaluation frameworks that better reflect the multilingual nature of our world. Future work expanding this approach to lower-resource languages and incorporating complementary notions of semantic plausibility will further illuminate the path toward truly inclusive AI systems that serve diverse linguistic communities more fairly.


\bibliography{wikidata-benchmark}

\begin{thebibliography}{38}
\providecommand{\natexlab}[1]{#1}

\bibitem[{AI(2024)}]{meta2024llama31}
Meta AI. 2024.
\newblock Llama 3.1 model card.
\newblock
  \url{https://github.com/meta-llama/llama-models/blob/main/models/llama3_1/MODEL_CARD.md}.
\newblock Accessed: 2026-02-06.

\bibitem[{Augenstein et~al.(2019)Augenstein, Lioma, Wang, Lima, Hansen, Hansen,
  and Simonsen}]{augenstein2019multifc}
Isabelle Augenstein, Christina Lioma, Dongsheng Wang, Lucas~Chaves Lima, Casper
  Hansen, Christian Hansen, and Jakob~Grue Simonsen. 2019.
\newblock Multifc: A real-world multi-domain dataset for evidence-based fact
  checking of claims.
\newblock In \emph{Proceedings of the 2019 Conference on Empirical Methods in
  Natural Language Processing and the 9th International Joint Conference on
  Natural Language Processing (EMNLP-IJCNLP)}, pages 4685--4697.

\bibitem[{Bayat et~al.(2025)Bayat, Zhang, Munir, and Wang}]{bayat2025factbench}
Farima~Fatahi Bayat, Lechen Zhang, Sheza Munir, and Lu~Wang. 2025.
\newblock Factbench: A dynamic benchmark for in-the-wild language model
  factuality evaluation.
\newblock In \emph{Proceedings of the 63rd Annual Meeting of the Association
  for Computational Linguistics (Volume 1: Long Papers)}, pages 33090--33110.

\bibitem[{Blasi et~al.(2022)Blasi, Anastasopoulos, and
  Neubig}]{blasi2022systematic}
Damian Blasi, Antonios Anastasopoulos, and Graham Neubig. 2022.
\newblock Systematic inequalities in language technology performance across the
  world’s languages.
\newblock In \emph{Proceedings of the 60th Annual Meeting of the Association
  for Computational Linguistics (Volume 1: Long Papers)}, pages 5486--5505.

\bibitem[{Cheng et~al.(2025)Cheng, Jacovi, Globerson, Golan, Kwong, Alberti,
  Tao, Ben-David, Tomar, Haas et~al.}]{cheng2025facts}
Aileen Cheng, Alon Jacovi, Amir Globerson, Ben Golan, Charles Kwong, Chris
  Alberti, Connie Tao, Eyal Ben-David, Gaurav~Singh Tomar, Lukas Haas, and 1
  others. 2025.
\newblock The facts leaderboard: A comprehensive benchmark for large language
  model factuality.
\newblock \emph{arXiv preprint arXiv:2512.10791}.

\bibitem[{Chollet et~al.(2025)Chollet, Knoop, Kamradt, Landers, and
  Pinkard}]{chollet2025arc}
Francois Chollet, Mike Knoop, Gregory Kamradt, Bryan Landers, and Henry
  Pinkard. 2025.
\newblock Arc-agi-2: A new challenge for frontier ai reasoning systems.
\newblock \emph{arXiv preprint arXiv:2505.11831}.

\bibitem[{Conneau et~al.(2018)Conneau, Rinott, Lample, Williams, Bowman,
  Schwenk, and Stoyanov}]{conneau2018xnli}
Alexis Conneau, Ruty Rinott, Guillaume Lample, Adina Williams, Samuel Bowman,
  Holger Schwenk, and Veselin Stoyanov. 2018.
\newblock Xnli: Evaluating cross-lingual sentence representations.
\newblock In \emph{Proceedings of the 2018 conference on empirical methods in
  natural language processing}, pages 2475--2485.

\bibitem[{Fierro and S{\o}gaard(2022)}]{fierro2022factual}
Constanza Fierro and Anders S{\o}gaard. 2022.
\newblock Factual consistency of multilingual pretrained language models.
\newblock In \emph{Findings of the Association for Computational Linguistics:
  ACL 2022}, pages 3046--3052.

\bibitem[{Gardner et~al.(2020)Gardner, Artzi, Basmov, Berant, Bogin, Chen,
  Dasigi, Dua, Elazar, Gottumukkala et~al.}]{gardner2020evaluating}
Matt Gardner, Yoav Artzi, Victoria Basmov, Jonathan Berant, Ben Bogin, Sihao
  Chen, Pradeep Dasigi, Dheeru Dua, Yanai Elazar, Ananth Gottumukkala, and 1
  others. 2020.
\newblock Evaluating models’ local decision boundaries via contrast sets.
\newblock In \emph{Findings of the Association for Computational Linguistics:
  EMNLP 2020}, pages 1307--1323.

\bibitem[{{Google DeepMind}(2025)}]{gemini3flash2025}
{Google DeepMind}. 2025.
\newblock Gemini 3 flash model card.
\newblock
  {\url{https://storage.googleapis.com/deepmind-media/Model-Cards/Gemini-3-Flash-Model-Card.pdf}}.
\newblock Accessed: 2026-02-06.

\bibitem[{Grover and Leskovec(2016)}]{grover2016node2vec}
Aditya Grover and Jure Leskovec. 2016.
\newblock node2vec: Scalable feature learning for networks.
\newblock In \emph{Proceedings of the 22nd ACM SIGKDD international conference
  on Knowledge discovery and data mining}, pages 855--864.

\bibitem[{Gupta et~al.(2025)Gupta, Mehta, Xu, and Srikumar}]{gupta2025found}
Ashim Gupta, Maitrey Mehta, Zhichao Xu, and Vivek Srikumar. 2025.
\newblock Found in translation: Measuring multilingual llm consistency as
  simple as translate then evaluate.
\newblock In \emph{Proceedings of the 14th International Joint Conference on
  Natural Language Processing and the 4th Conference of the Asia-Pacific
  Chapter of the Association for Computational Linguistics}, pages 3477--3496.

\bibitem[{Gupta and Srikumar(2021)}]{gupta2021x}
Ashim Gupta and Vivek Srikumar. 2021.
\newblock X-fact: A new benchmark dataset for multilingual fact checking.
\newblock In \emph{Proceedings of the 59th Annual Meeting of the Association
  for Computational Linguistics and the 11th International Joint Conference on
  Natural Language Processing (Volume 2: Short Papers)}, pages 675--682.

\bibitem[{Hendrycks et~al.(2021)Hendrycks, Burns, Basart, Zou, Mazeika, Song,
  and Steinhardt}]{hendrycks2021measuring}
Dan Hendrycks, Collin Burns, Steven Basart, Andy Zou, Mantas Mazeika, Dawn
  Song, and Jacob Steinhardt. 2021.
\newblock \href {https://openreview.net/forum?id=d7KBjmI3GmQ} {Measuring
  massive multitask language understanding}.
\newblock In \emph{Proceedings of the International Conference on Learning
  Representations (ICLR)}.

\bibitem[{Hu et~al.(2020)Hu, Ruder, Siddhant, Neubig, Firat, and
  Johnson}]{hu2020xtreme}
Junjie Hu, Sebastian Ruder, Aditya Siddhant, Graham Neubig, Orhan Firat, and
  Melvin Johnson. 2020.
\newblock Xtreme: A massively multilingual multi-task benchmark for evaluating
  cross-lingual generalisation.
\newblock In \emph{International conference on machine learning}, pages
  4411--4421. PMLR.

\bibitem[{Kassner et~al.(2021)Kassner, Dufter, and
  Sch{\"u}tze}]{kassner2021multilingual}
Nora Kassner, Philipp Dufter, and Hinrich Sch{\"u}tze. 2021.
\newblock Multilingual lama: Investigating knowledge in multilingual pretrained
  language models.
\newblock In \emph{Proceedings of the 16th Conference of the European Chapter
  of the Association for Computational Linguistics: Main Volume}, pages
  3250--3258.

\bibitem[{{Mistral AI Team}(2024)}]{mistral_nemo_2024}
{Mistral AI Team}. 2024.
\newblock Mistral-nemo-instruct-2407.
\newblock Hugging Face,
  \url{https://huggingface.co/mistralai/Mistral-Nemo-Instruct-2407}.
\newblock Accessed: 2026-02-06.

\bibitem[{Mousavi et~al.(2024)Mousavi, Alghisi, and
  Riccardi}]{mousavi2024dyknow}
Seyed~Mahed Mousavi, Simone Alghisi, and Giuseppe Riccardi. 2024.
\newblock Dyknow: Dynamically verifying time-sensitive factual knowledge in
  llms.
\newblock In \emph{Findings of the Association for Computational Linguistics:
  EMNLP 2024}, pages 8014--8029.

\bibitem[{OpenAI(2024)}]{openai2024mmmlu}
OpenAI. 2024.
\newblock Multilingual massive multitask language understanding.
\newblock \url{https://huggingface.co/datasets/openai/MMMLU}.
\newblock Professional human translations into 14 languages.

\bibitem[{Panickssery et~al.(2024)Panickssery, Bowman, and
  Feng}]{panickssery2024llm}
Arjun Panickssery, Samuel Bowman, and Shi Feng. 2024.
\newblock Llm evaluators recognize and favor their own generations.
\newblock \emph{Advances in Neural Information Processing Systems},
  37:68772--68802.

\bibitem[{Perozzi et~al.(2014)Perozzi, Al-Rfou, and
  Skiena}]{perozzi2014deepwalk}
Bryan Perozzi, Rami Al-Rfou, and Steven Skiena. 2014.
\newblock Deepwalk: Online learning of social representations.
\newblock In \emph{Proceedings of the 20th ACM SIGKDD international conference
  on Knowledge discovery and data mining}, pages 701--710.

\bibitem[{Phan et~al.(2025)Phan, Gatti, Han, Li, Hu, Zhang, Zhang, Shaaban,
  Ling, Shi et~al.}]{phan2025humanity}
Long Phan, Alice Gatti, Ziwen Han, Nathaniel Li, Josephina Hu, Hugh Zhang, Chen
  Bo~Calvin Zhang, Mohamed Shaaban, John Ling, Sean Shi, and 1 others. 2025.
\newblock Humanity's last exam.
\newblock \emph{arXiv preprint arXiv:2501.14249}.

\bibitem[{Qi et~al.(2023)Qi, Fern{\'a}ndez, and Bisazza}]{qi2023cross}
Jirui Qi, Raquel Fern{\'a}ndez, and Arianna Bisazza. 2023.
\newblock Cross-lingual consistency of factual knowledge in multilingual
  language models.
\newblock In \emph{Proceedings of the 2023 Conference on Empirical Methods in
  Natural Language Processing}, pages 10650--10666.

\bibitem[{Rajaee and Monz(2024)}]{rajaee2024analyzing}
Sara Rajaee and Christof Monz. 2024.
\newblock Analyzing the evaluation of cross-lingual knowledge transfer in
  multilingual language models.
\newblock In \emph{Proceedings of the 18th Conference of the European Chapter
  of the Association for Computational Linguistics (Volume 1: Long Papers)},
  pages 2895--2914.

\bibitem[{Rein et~al.(2024)Rein, Hou, Stickland, Petty, Pang, Dirani, Michael,
  and Bowman}]{rein2024gpqa}
David Rein, Betty~Li Hou, Asa~Cooper Stickland, Jackson Petty, Richard~Yuanzhe
  Pang, Julien Dirani, Julian Michael, and Samuel~R Bowman. 2024.
\newblock Gpqa: A graduate-level google-proof q\&a benchmark.
\newblock In \emph{First Conference on Language Modeling}.

\bibitem[{Schut et~al.(2025)Schut, Gal, and Farquhar}]{schut2025multilingual}
Lisa Schut, Yarin Gal, and Sebastian Farquhar. 2025.
\newblock Do multilingual llms think in english?
\newblock \emph{arXiv preprint arXiv:2502.15603}.

\bibitem[{Shen et~al.(2024)Shen, Tan, Chen, Chen, Zhang, Xu, Zheng, Koehn, and
  Khashabi}]{shen2024language}
Lingfeng Shen, Weiting Tan, Sihao Chen, Yunmo Chen, Jingyu Zhang, Haoran Xu,
  Boyuan Zheng, Philipp Koehn, and Daniel Khashabi. 2024.
\newblock The language barrier: Dissecting safety challenges of llms in
  multilingual contexts.
\newblock In \emph{Findings of the Association for Computational Linguistics
  ACL 2024}, pages 2668--2680.

\bibitem[{Singh et~al.(2025)Singh, Romanou, Fourrier, Adelani, Ngui,
  Vila-Suero, Limkonchotiwat, Marchisio, Leong, Susanto
  et~al.}]{singh2025global}
Shivalika Singh, Angelika Romanou, Cl{\'e}mentine Fourrier, David~Ifeoluwa
  Adelani, Jian~Gang Ngui, Daniel Vila-Suero, Peerat Limkonchotiwat, Kelly
  Marchisio, Wei~Qi Leong, Yosephine Susanto, and 1 others. 2025.
\newblock Global mmlu: Understanding and addressing cultural and linguistic
  biases in multilingual evaluation.
\newblock In \emph{Proceedings of the 63rd Annual Meeting of the Association
  for Computational Linguistics (Volume 1: Long Papers)}, pages 18761--18799.

\bibitem[{StarscreamDeceptions(2025)}]{multilingual_mmlu_leaderboard}
StarscreamDeceptions. 2025.
\newblock Multilingual mmlu benchmark leaderboard.
\newblock Hugging Face Space,
  \url{https://huggingface.co/spaces/StarscreamDeceptions/Multilingual-MMLU-Benchmark-Leaderboard}.
\newblock Accessed: 2026-02-06.

\bibitem[{Ul~Islam et~al.(2025)Ul~Islam, Lauscher, and
  Glava{\v{s}}}]{ulislam2025hallucinate}
Saad~Obaid Ul~Islam, Anne Lauscher, and Goran Glava{\v{s}}. 2025.
\newblock How much do {LLM}s hallucinate across languages? on realistic
  multilingual estimation of {LLM} hallucination.
\newblock In \emph{Proceedings of the 2025 Conference on Empirical Methods in
  Natural Language Processing}, pages 29077--29098.

\bibitem[{Veseli et~al.(2023)Veseli, Singhania, Razniewski, and
  Weikum}]{veseli2023evaluating}
Blerta Veseli, Sneha Singhania, Simon Razniewski, and Gerhard Weikum. 2023.
\newblock Evaluating language models for knowledge base completion.
\newblock In \emph{European Semantic Web Conference}, pages 227--243. Springer.

\bibitem[{Vrande{\v{c}}i{\'c} and Kr{\"o}tzsch(2014)}]{vrandevcic2014wikidata}
Denny Vrande{\v{c}}i{\'c} and Markus Kr{\"o}tzsch. 2014.
\newblock Wikidata: a free collaborative knowledgebase.
\newblock \emph{Communications of the ACM}, 57(10):78--85.

\bibitem[{Wang et~al.(2019)Wang, Pruksachatkun, Nangia, Singh, Michael, Hill,
  Levy, and Bowman}]{wang2019superglue}
Alex Wang, Yada Pruksachatkun, Nikita Nangia, Amanpreet Singh, Julian Michael,
  Felix Hill, Omer Levy, and Samuel Bowman. 2019.
\newblock Superglue: A stickier benchmark for general-purpose language
  understanding systems.
\newblock \emph{Advances in neural information processing systems}, 32.

\bibitem[{Wang et~al.(2018)Wang, Singh, Michael, Hill, Levy, and
  Bowman}]{wang2018glue}
Alex Wang, Amanpreet Singh, Julian Michael, Felix Hill, Omer Levy, and Samuel~R
  Bowman. 2018.
\newblock Glue: A multi-task benchmark and analysis platform for natural
  language understanding.
\newblock In \emph{Proceedings of the 2018 EMNLP workshop BlackboxNLP:
  Analyzing and interpreting neural networks for NLP}, pages 353--355.

\bibitem[{Wang et~al.(2025)Wang, Huang, Yang, Xie, and
  Kawahara}]{wang2025traveling}
Hao Wang, Pinzhi Huang, Jihan Yang, Saining Xie, and Daisuke Kawahara. 2025.
\newblock Traveling across languages: Benchmarking cross-lingual consistency in
  multimodal llms.
\newblock \emph{arXiv preprint arXiv:2505.15075}.

\bibitem[{Wang et~al.(2024)Wang, Ma, Zhang, Ni, Chandra, Guo, Ren, Arulraj, He,
  Jiang et~al.}]{wang2024mmlu}
Yubo Wang, Xueguang Ma, Ge~Zhang, Yuansheng Ni, Abhranil Chandra, Shiguang Guo,
  Weiming Ren, Aaran Arulraj, Xuan He, Ziyan Jiang, and 1 others. 2024.
\newblock Mmlu-pro: A more robust and challenging multi-task language
  understanding benchmark.
\newblock \emph{Advances in Neural Information Processing Systems},
  37:95266--95290.

\bibitem[{Yang et~al.(2025)Yang, Li, Yang, Zhang, Hui, Zheng, Yu, Gao, Huang,
  Lv et~al.}]{yang2025qwen3}
An~Yang, Anfeng Li, Baosong Yang, Beichen Zhang, Binyuan Hui, Bo~Zheng, Bowen
  Yu, Chang Gao, Chengen Huang, Chenxu Lv, and 1 others. 2025.
\newblock Qwen3 technical report.
\newblock \emph{arXiv preprint arXiv:2505.09388}.

\bibitem[{Zhang et~al.(2025)Zhang, Anjum, Fan, Zheng, Huang, and
  Feng}]{zhang2025polyfever}
Hanzhi Zhang, Sumera Anjum, Heng Fan, Weijian Zheng, Yan Huang, and Yunhe Feng.
  2025.
\newblock Poly-fever: A multilingual fact verification benchmark for
  hallucination detection in large language models.
\newblock \emph{arXiv preprint arXiv:2503.16541}.

\end{thebibliography}


\clearpage
\appendix

\section{AI Disclosure}
AI assistants (Claude and Gemini) were used for language polishing and coding assistance in this work. These tools were used to refine the writing of the paper and assist with implementation code. All methods, results, and conclusions are the sole responsibility of the authors.

\section{Prompt Templates}\label{sec:prompts}

The following prompts are used for multilingual statement generation (Section~\ref{subsec:generation}) and true/false classification evaluation (Section~\ref{subsec:eval}), respectively. The placeholders in the generation prompt are filled with the extracted Wikidata labels for the subject, property, and object in the target language. Also, the resulting sentence generated by Gemini-2.5-Flash with default mode is then used to fill the \texttt{\{Testing statement\}} placeholder in the evaluation prompt.

\begin{promptbox}{Generation Prompt}
You are a \textbf{\{target language\}} language writing expert. Write one concise and grammatically correct sentence in \textbf{\{target language\}} that expresses the given relation without adding or modifying any facts. Respond only in \textbf{\{target language\}} and enclose your answer within <answer> and </answer> tags. Do not include explanations, examples, or any text outside the tags. For example: <answer>Washington, D.C. is the capital of the United States.</answer> Now generate one sentence for this input triple: [\textbf{\{Subject in target language\}}, \textbf{\{Property in target language\}}, \textbf{\{Object in target language\}}]
\end{promptbox} 

\begin{promptbox}{Evaluation Prompt}
Is the following statement True or False? statement: \textbf{\{Testing statement\}}. Answer must be one of the following options: True, False, Unsure. Answer must be encapsulated with <answer></answer>
\end{promptbox} 

\section{Computational Setup}
Statement generation was performed using the Gemini-2.5-Flash API. Local model evaluation (LLaMA 3.1, Mistral Nemo, Qwen3) was conducted on a system equipped with $6\times$ NVIDIA GeForce RTX 4090 GPUs (24GB VRAM each), dual Intel Xeon Gold 6326 CPUs, and 1TB DDR4 ECC memory using VLLM. Knowledge graph embedding computation (node2vec) used CPU resources in the same system, implemented with Gensim's \texttt{models.word2vec}. 

\section{Appendix Tables and Figures}

\begin{table*}[h]
\centering
\caption{Multilingual MMLU accuracy (\%) across models. Scores are primarily from official model cards, supplemented by leaderboard results~\cite{multilingual_mmlu_leaderboard} for languages not covered in official sources (notably Korean for LLaMA-3.1 and Mistral-Nemo, and Japanese for LLaMA-3.1). Note that LLaMA-3.1 does not officially support Korean or Japanese, and Mistral-Nemo claims Korean support but does not report Korean performance in its official model card. Reported averages are aggregate scores computed over the languages available for each model as reported by the source. For Qwen models, numbers outside parentheses are with reasoning (think mode), while numbers in parentheses are without reasoning (non-think mode). Note that English MMLU results for Qwen post-trained models are not available, and thus, we report results from pre-trained models, and MMMLU scores for Gemini-2.5-Flash are only reported in the Gemini 3 Flash model card, instead of Gemini-2.5-Flash model card.}\label{tab:mmlu_multilingual}
\resizebox{\textwidth}{!}{
\begin{tabular}{ccccccc}
\toprule
\textbf{Language} &
\textbf{\makecell{Mistral-Nemo\\-Instruct-2407 \cite{mistral_nemo_2024}}} &
\textbf{\makecell{LLaMA-3.1-8B \\ (Model Card \cite{meta2024llama31})}} &
\textbf{\makecell{LLaMA-3.1-8B \\ (Leaderboard \cite{multilingual_mmlu_leaderboard})}} &
\textbf{\makecell{Gemini-2.5\\-Flash \cite{gemini3flash2025}}} &
\textbf{Qwen3-8B \cite{yang2025qwen3}}&
\textbf{Qwen3-14B \cite{yang2025qwen3}} \\
\midrule
Reported Avg. & 61.42 & 58.62 & 50.01 & 86.6 & 74.4 (66.9) & 77.9 (72.6) \\
\midrule
EN & 68.0 & 69.4 & -- & -- & 76.89 (Pretraining) & 81.05 (Pretraining) \\
DE & 62.7 & 60.59  & 55.63 & -- & 78.1 (70.8) & 80.9 (76.0) \\
FR & 62.3 & 62.34 & 58.94 & -- & 77.9 (71.7) & 80.4 (75.1) \\
ES & 64.6 & 62.45 & 59.14 & -- & 79.2 (73.7) & 81.1 (77.4) \\
IT & 61.3 & 61.63 & 56.27 & -- & 77.5 (72.9) & 80.2 (75.7) \\
PT & 63.3 & 62.12 & 59.02 & -- & 78.6 (72.9) & 81.6 (76.5) \\
KO & -- & -- & 50.78 & -- & 74.7 (66.5) & 79.6 (73.3) \\
JA & 59.0 & -- & 52.07 & -- & 74.9 (69.9) & 79.4 (73.8) \\
\bottomrule
\end{tabular}
}
\end{table*}

\begin{figure*}[hb]
    \centering
    \includegraphics[width=\textwidth]{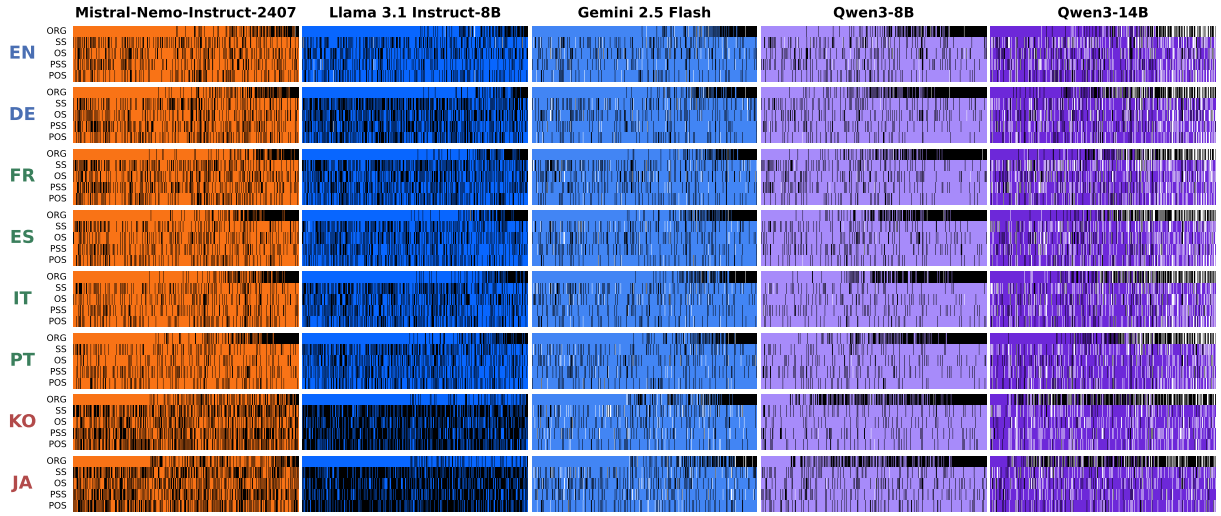}
    \caption{Statement-level response visualization sorted by difficulty. Columns represent 500 statements sorted by ORG accuracy; rows show language-distortion combinations. Colored cells (orange/blues/purples) are correct, black cells are incorrect, and white cells are unsure or invalid responses. Western languages exhibit predominantly correct responses (colored) with gradual degradation. East Asian languages in LLaMA and Mistral show extensive incorrect (black) responses, particularly for harder problems. Gemini and Qwen maintain high correctness (colored patterns) across all languages. White regions reveal model uncertainty that aggregate metrics cannot capture. LLaMA and Mistral show virtually zero \texttt{Unsure} responses (5-run avg. 0.47 and 0.03 out of 500, respectively), confirming that their cross-lingual gaps are driven entirely by incorrect responses rather than abstention. Qwen3 models (non-think mode) show higher \texttt{Unsure} rates (Qwen3-8B: avg. 34.17 out of 500, 6.8\%) but distributed consistently across languages, introducing no systematic cross-lingual bias.} 
    \label{fig:heatmap_spectrum}
\end{figure*}

\begin{figure*}[hb]
    \centering
    \includegraphics[width=\textwidth]{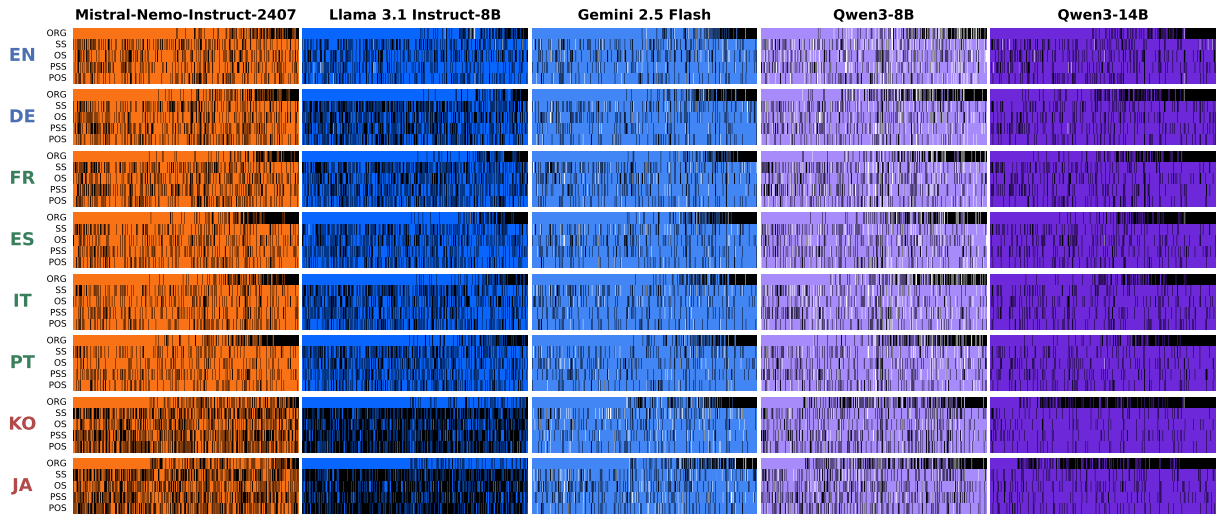}
    \caption{Statement-level response visualization, constructed as in Figure~\ref{fig:heatmap_spectrum}. Only Qwen3 models are replaced with think mode results. Compared with non-think mode, \texttt{Unsure} responses decrease visibly for Qwen3-8B but increase for Qwen3-14B.}
    \label{fig:heatmap_spectrum_think}
\end{figure*}

\begin{figure*}[h]
    \centering
    \includegraphics[width=\textwidth]{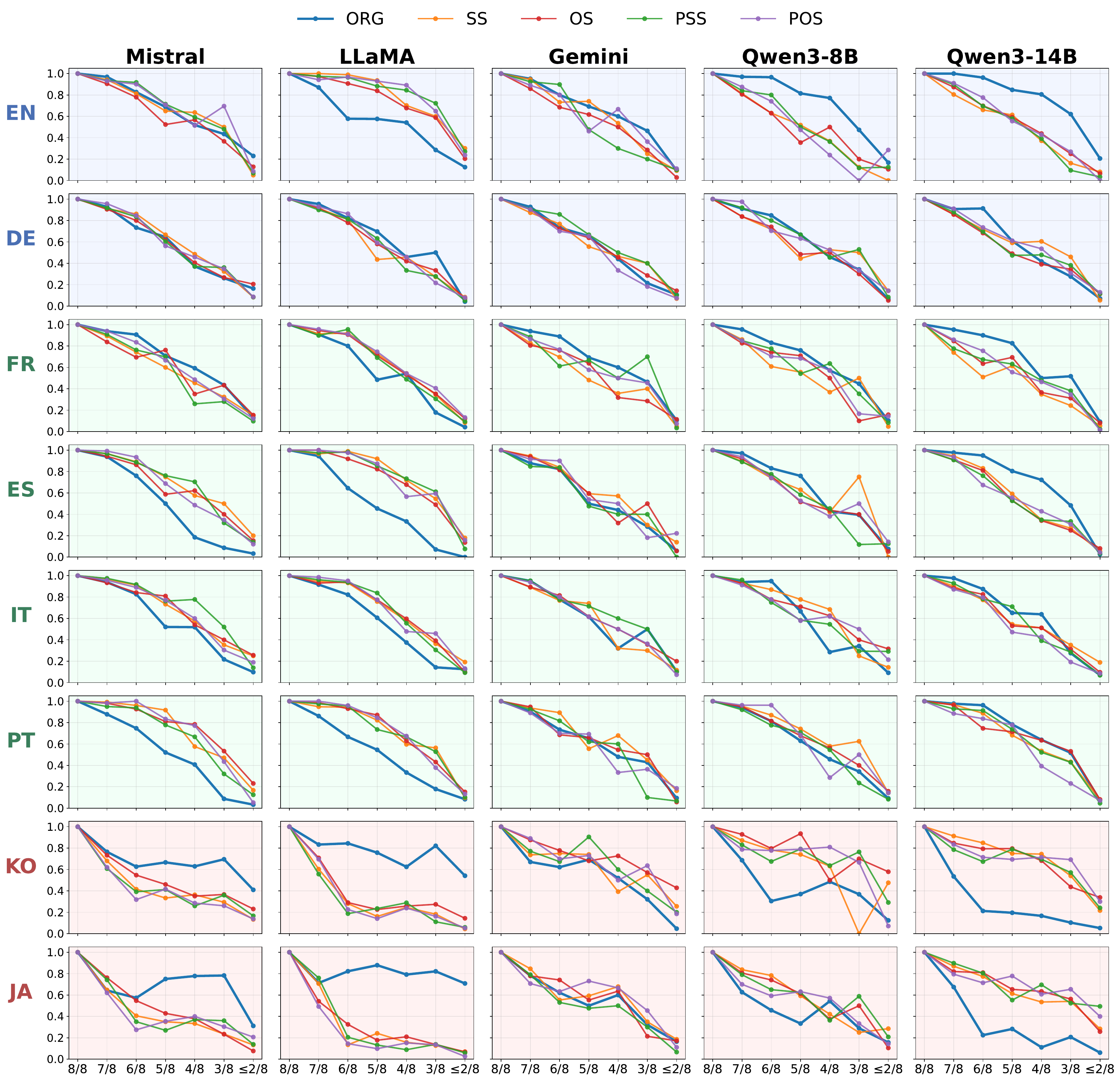}
    \caption{Accuracy trends across problem difficulty buckets (alternative visualization of Figure~\ref{fig:heatmap} to supplement, with Qwen3 non-think mode). Each subplot shows accuracy changes as problem difficulty decreases (x-axis: 8/8 = easiest to $\leq$2/8 = hardest; y-axis: accuracy) for one model-language combination. Five lines represent distortion types: Original (ORG), Subject Shuffle (SS), Object Shuffle (OS), Property-based Subject Shuffle (PSS), and Property-based Object Shuffle (POS). Western languages (blue and green background) exhibit smooth, consistent degradation across difficulty levels, while East Asian languages (coral background) show sudden drops as difficulty increases, particularly in LLaMA and Mistral. This visualization reveals language-specific instabilities that aggregate accuracy metrics may obscure.}\label{fig:heatmap_line}
\end{figure*}

\begin{figure*}[h]
    \centering
    \includegraphics[width=\textwidth]{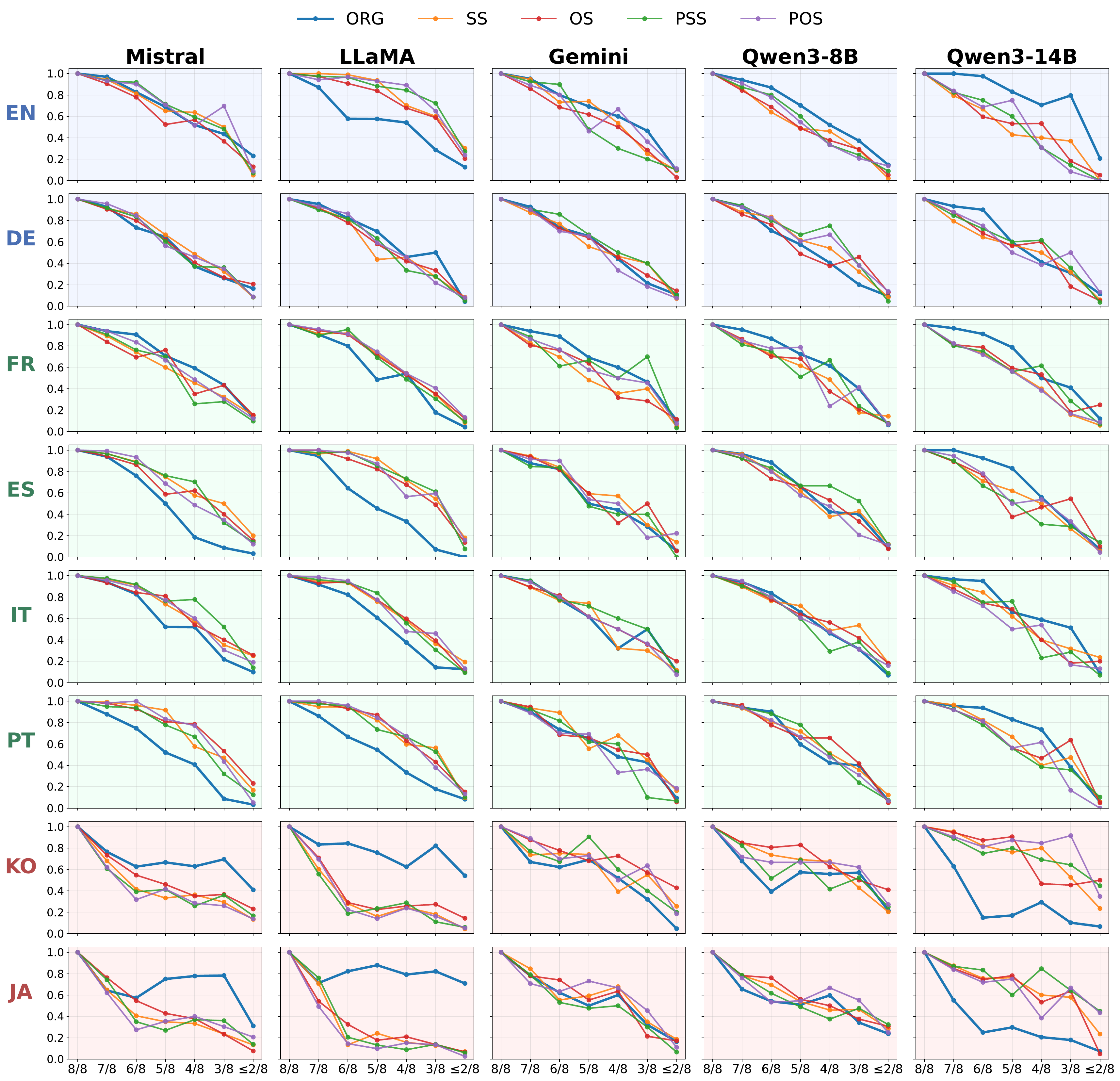}
    \caption{Accuracy trends across problem difficulty buckets, constructed as in Figure~\ref{fig:heatmap_line}. Only Qwen3 models are replaced with think-mode results. The overall trend remains similar to the non-think-mode results, with Asian-language degradation still most visible in LLaMA and Mistral.}
    \label{fig:heatmap_line_think}
\end{figure*}

\begin{table*}[t]
\centering
\caption{Accuracy with Unsure responses counted as incorrect}
\label{tab:accuracy_by_lang_with_unsure}
\vspace{-3mm}
\scriptsize
\setlength{\tabcolsep}{2.4pt}
\renewcommand{\arraystretch}{0.95}
\resizebox{\textwidth}{!}{%
\begin{tabular}{llcccccccc}
\toprule
\textbf{Model} & \textbf{Type} & \textbf{EN} & \textbf{DE} & \textbf{FR} & \textbf{ES} & \textbf{IT} & \textbf{PT} & \textbf{KO} & \textbf{JA} \\
\midrule
\multirow{5}{*}{\makecell{Mistral\\-Nemo\\-Instruct\\-2407}} & ORG & \textbf{0.792$\pm$0.000} & 0.742$\pm$0.001 & \textbf{0.792$\pm$0.000} & 0.700$\pm$0.000 & 0.744$\pm$0.001 & 0.700$\pm$0.000 & 0.760$\pm$0.000 & 0.736$\pm$0.001 \\
 & SS & 0.734$\pm$0.001 & 0.720$\pm$0.000 & 0.690$\pm$0.000 & 0.782$\pm$0.000 & 0.780$\pm$0.000 & \textbf{0.816$\pm$0.001} & 0.530$\pm$0.000 & 0.517$\pm$0.001 \\
 & OS & 0.740$\pm$0.001 & 0.744$\pm$0.000 & 0.731$\pm$0.002 & 0.780$\pm$0.000 & 0.803$\pm$0.002 & \textbf{0.854$\pm$0.000} & 0.643$\pm$0.001 & 0.628$\pm$0.000 \\
 & PSS & 0.750$\pm$0.000 & 0.698$\pm$0.001 & 0.686$\pm$0.001 & 0.768$\pm$0.001 & \textbf{0.787$\pm$0.001} & 0.770$\pm$0.000 & 0.522$\pm$0.001 & 0.530$\pm$0.000 \\
 & POS & 0.784$\pm$0.000 & 0.746$\pm$0.001 & 0.755$\pm$0.002 & 0.789$\pm$0.002 & 0.794$\pm$0.000 & \textbf{0.828$\pm$0.001} & 0.548$\pm$0.000 & 0.552$\pm$0.000 \\
\midrule
\multirow{5}{*}{\makecell{LLaMA\\-3.1\\-Instruct\\-8B}} & ORG & 0.801$\pm$0.001 & 0.854$\pm$0.000 & 0.814$\pm$0.000 & 0.788$\pm$0.000 & 0.820$\pm$0.001 & 0.788$\pm$0.000 & \textbf{0.884$\pm$0.000} & 0.880$\pm$0.001 \\
 & SS & \textbf{0.695$\pm$0.001} & 0.458$\pm$0.001 & 0.532$\pm$0.000 & 0.646$\pm$0.000 & 0.580$\pm$0.002 & 0.586$\pm$0.001 & 0.248$\pm$0.000 & 0.237$\pm$0.001 \\
 & OS & \textbf{0.668$\pm$0.000} & 0.512$\pm$0.000 & 0.584$\pm$0.000 & 0.644$\pm$0.001 & 0.602$\pm$0.001 & 0.640$\pm$0.001 & 0.348$\pm$0.000 & 0.284$\pm$0.001 \\
 & PSS & \textbf{0.767$\pm$0.001} & 0.556$\pm$0.000 & 0.622$\pm$0.000 & 0.702$\pm$0.000 & 0.654$\pm$0.000 & 0.674$\pm$0.000 & 0.294$\pm$0.001 & 0.298$\pm$0.002 \\
 & POS & \textbf{0.764$\pm$0.000} & 0.580$\pm$0.001 & 0.654$\pm$0.001 & 0.714$\pm$0.001 & 0.668$\pm$0.000 & 0.694$\pm$0.001 & 0.298$\pm$0.002 & 0.226$\pm$0.000 \\
\midrule
\multirow{5}{*}{\makecell{Gemini\\-2.5\\-Flash}} & ORG & 0.754 & 0.722 & \textbf{0.762} & 0.710 & 0.738 & 0.730 & 0.672 & 0.704 \\
 & SS & 0.810 & 0.788 & 0.758 & 0.822 & 0.794 & \textbf{0.838} & 0.784 & 0.780 \\
 & OS & 0.788 & 0.810 & 0.784 & 0.816 & 0.824 & 0.822 & \textbf{0.854} & 0.784 \\
 & PSS & 0.868 & 0.876 & 0.850 & 0.846 & \textbf{0.884} & 0.868 & 0.848 & 0.808 \\
 & POS & 0.876 & 0.872 & 0.874 & 0.890 & 0.888 & 0.882 & \textbf{0.892} & 0.852 \\
\midrule
\multirow{5}{*}{\makecell{Qwen3\\-8B\\Non-think}} & ORG & \textbf{0.608$\pm$0.000} & 0.500$\pm$0.001 & 0.543$\pm$0.001 & 0.521$\pm$0.001 & 0.513$\pm$0.001 & 0.505$\pm$0.001 & 0.401$\pm$0.001 & 0.416$\pm$0.000 \\
 & SS & 0.828$\pm$0.000 & 0.853$\pm$0.001 & 0.844$\pm$0.000 & 0.871$\pm$0.002 & 0.904$\pm$0.000 & \textbf{0.908$\pm$0.000} & 0.892$\pm$0.001 & 0.866$\pm$0.000 \\
 & OS & 0.817$\pm$0.001 & 0.842$\pm$0.000 & 0.854$\pm$0.000 & 0.862$\pm$0.000 & 0.894$\pm$0.000 & 0.892$\pm$0.000 & \textbf{0.921$\pm$0.001} & 0.847$\pm$0.001 \\
 & PSS & 0.842$\pm$0.000 & 0.880$\pm$0.000 & 0.858$\pm$0.001 & 0.856$\pm$0.001 & \textbf{0.884$\pm$0.000} & 0.872$\pm$0.000 & 0.880$\pm$0.000 & 0.846$\pm$0.000 \\
 & POS & 0.882$\pm$0.000 & 0.915$\pm$0.001 & 0.896$\pm$0.000 & 0.902$\pm$0.000 & 0.914$\pm$0.000 & \textbf{0.920$\pm$0.000} & 0.908$\pm$0.000 & 0.866$\pm$0.000 \\
\midrule
\multirow{5}{*}{\makecell{Qwen3\\-14B\\Non-think}} & ORG & \textbf{0.580$\pm$0.001} & 0.434$\pm$0.001 & 0.486$\pm$0.000 & 0.478$\pm$0.000 & 0.454$\pm$0.001 & 0.494$\pm$0.001 & 0.216$\pm$0.001 & 0.245$\pm$0.001 \\
 & SS & 0.641$\pm$0.001 & 0.695$\pm$0.001 & 0.614$\pm$0.001 & 0.688$\pm$0.001 & 0.708$\pm$0.000 & 0.735$\pm$0.001 & \textbf{0.774$\pm$0.001} & 0.739$\pm$0.001 \\
 & OS & 0.683$\pm$0.001 & 0.676$\pm$0.001 & 0.677$\pm$0.002 & 0.694$\pm$0.001 & 0.710$\pm$0.000 & 0.758$\pm$0.000 & \textbf{0.776$\pm$0.001} & 0.752$\pm$0.000 \\
 & PSS & 0.686$\pm$0.000 & 0.704$\pm$0.000 & 0.677$\pm$0.001 & 0.698$\pm$0.001 & 0.722$\pm$0.001 & 0.746$\pm$0.000 & 0.748$\pm$0.000 & \textbf{0.806$\pm$0.000} \\
 & POS & 0.722$\pm$0.000 & 0.747$\pm$0.001 & 0.720$\pm$0.000 & 0.724$\pm$0.000 & 0.720$\pm$0.001 & 0.746$\pm$0.000 & 0.794$\pm$0.000 & \textbf{0.799$\pm$0.002} \\
\bottomrule
\end{tabular}
}
\vspace{-4mm}
\end{table*}
\begin{table*}[t]
\centering
\caption{Accuracy with Unsure responses excluded from scoring}
\label{tab:accuracy_by_lang_wo_unsure}
\vspace{-3mm}
\scriptsize
\setlength{\tabcolsep}{2.4pt}
\renewcommand{\arraystretch}{0.95}
\resizebox{\textwidth}{!}{%
\begin{tabular}{llcccccccc}
\toprule
\textbf{Model} & \textbf{Type} & \textbf{EN} & \textbf{DE} & \textbf{FR} & \textbf{ES} & \textbf{IT} & \textbf{PT} & \textbf{KO} & \textbf{JA} \\
\midrule
\multirow{5}{*}{\makecell{Mistral\\-Nemo\\-Instruct\\-2407}} & ORG & \textbf{0.794$\pm$0.000} & 0.742$\pm$0.001 & 0.792$\pm$0.000 & 0.700$\pm$0.000 & 0.744$\pm$0.001 & 0.700$\pm$0.000 & 0.760$\pm$0.000 & 0.736$\pm$0.001 \\
 & SS & 0.734$\pm$0.001 & 0.720$\pm$0.000 & 0.690$\pm$0.000 & 0.782$\pm$0.000 & 0.780$\pm$0.000 & \textbf{0.816$\pm$0.001} & 0.530$\pm$0.000 & 0.517$\pm$0.001 \\
 & OS & 0.740$\pm$0.001 & 0.744$\pm$0.000 & 0.731$\pm$0.002 & 0.780$\pm$0.000 & 0.803$\pm$0.002 & \textbf{0.854$\pm$0.000} & 0.643$\pm$0.001 & 0.628$\pm$0.000 \\
 & PSS & 0.750$\pm$0.000 & 0.698$\pm$0.001 & 0.686$\pm$0.001 & 0.768$\pm$0.001 & \textbf{0.787$\pm$0.001} & 0.770$\pm$0.000 & 0.522$\pm$0.001 & 0.530$\pm$0.000 \\
 & POS & 0.784$\pm$0.000 & 0.746$\pm$0.001 & 0.755$\pm$0.002 & 0.789$\pm$0.002 & 0.794$\pm$0.000 & \textbf{0.828$\pm$0.001} & 0.548$\pm$0.000 & 0.552$\pm$0.000 \\
\midrule
\multirow{5}{*}{\makecell{LLaMA\\-3.1\\-Instruct\\-8B}} & ORG & 0.808$\pm$0.001 & 0.854$\pm$0.000 & 0.814$\pm$0.000 & 0.790$\pm$0.000 & 0.820$\pm$0.001 & 0.788$\pm$0.000 & \textbf{0.884$\pm$0.000} & 0.880$\pm$0.001 \\
 & SS & \textbf{0.701$\pm$0.001} & 0.458$\pm$0.001 & 0.532$\pm$0.000 & 0.647$\pm$0.000 & 0.580$\pm$0.002 & 0.586$\pm$0.001 & 0.248$\pm$0.000 & 0.237$\pm$0.001 \\
 & OS & \textbf{0.668$\pm$0.000} & 0.512$\pm$0.000 & 0.584$\pm$0.000 & 0.644$\pm$0.001 & 0.602$\pm$0.001 & 0.640$\pm$0.001 & 0.348$\pm$0.000 & 0.284$\pm$0.001 \\
 & PSS & \textbf{0.775$\pm$0.001} & 0.557$\pm$0.000 & 0.622$\pm$0.000 & 0.702$\pm$0.000 & 0.654$\pm$0.000 & 0.674$\pm$0.000 & 0.294$\pm$0.001 & 0.298$\pm$0.002 \\
 & POS & \textbf{0.769$\pm$0.000} & 0.580$\pm$0.001 & 0.654$\pm$0.001 & 0.714$\pm$0.001 & 0.668$\pm$0.000 & 0.694$\pm$0.001 & 0.298$\pm$0.002 & 0.226$\pm$0.000 \\
\midrule
\multirow{5}{*}{\makecell{Gemini\\-2.5\\-Flash}} & ORG & 0.754 & 0.722 & \textbf{0.762} & 0.710 & 0.738 & 0.730 & 0.672 & 0.704 \\
 & SS & 0.810 & 0.788 & 0.758 & 0.822 & 0.794 & \textbf{0.838} & 0.784 & 0.780 \\
 & OS & 0.788 & 0.810 & 0.784 & 0.816 & 0.824 & 0.822 & \textbf{0.854} & 0.784 \\
 & PSS & 0.868 & 0.876 & 0.850 & 0.846 & \textbf{0.884} & 0.868 & 0.848 & 0.808 \\
 & POS & 0.876 & 0.872 & 0.874 & 0.890 & 0.888 & 0.882 & \textbf{0.892} & 0.852 \\
\midrule
\multirow{5}{*}{\makecell{Qwen3\\-8B\\Non-think}} & ORG & \textbf{0.613$\pm$0.000} & 0.503$\pm$0.001 & 0.545$\pm$0.001 & 0.523$\pm$0.001 & 0.516$\pm$0.001 & 0.505$\pm$0.001 & 0.402$\pm$0.001 & 0.416$\pm$0.000 \\
 & SS & 0.845$\pm$0.000 & 0.855$\pm$0.001 & 0.847$\pm$0.000 & 0.876$\pm$0.002 & \textbf{0.909$\pm$0.000} & 0.908$\pm$0.000 & 0.892$\pm$0.001 & 0.869$\pm$0.000 \\
 & OS & 0.827$\pm$0.001 & 0.844$\pm$0.000 & 0.857$\pm$0.000 & 0.864$\pm$0.000 & 0.898$\pm$0.000 & 0.894$\pm$0.000 & \textbf{0.921$\pm$0.001} & 0.850$\pm$0.001 \\
 & PSS & 0.852$\pm$0.000 & 0.882$\pm$0.000 & 0.863$\pm$0.001 & 0.858$\pm$0.001 & \textbf{0.886$\pm$0.000} & 0.874$\pm$0.000 & 0.884$\pm$0.000 & 0.849$\pm$0.000 \\
 & POS & 0.886$\pm$0.000 & 0.917$\pm$0.001 & 0.896$\pm$0.000 & 0.904$\pm$0.000 & 0.914$\pm$0.000 & \textbf{0.920$\pm$0.000} & 0.910$\pm$0.000 & 0.866$\pm$0.000 \\
\midrule
\multirow{5}{*}{\makecell{Qwen3\\-14B\\Non-think}} & ORG & \textbf{0.718$\pm$0.001} & 0.534$\pm$0.001 & 0.612$\pm$0.000 & 0.593$\pm$0.000 & 0.572$\pm$0.001 & 0.603$\pm$0.001 & 0.274$\pm$0.001 & 0.301$\pm$0.001 \\
 & SS & 0.853$\pm$0.000 & 0.868$\pm$0.001 & 0.855$\pm$0.000 & 0.893$\pm$0.000 & 0.905$\pm$0.000 & 0.922$\pm$0.000 & \textbf{0.942$\pm$0.001} & 0.911$\pm$0.000 \\
 & OS & 0.830$\pm$0.001 & 0.837$\pm$0.000 & 0.852$\pm$0.000 & 0.863$\pm$0.000 & 0.877$\pm$0.000 & 0.890$\pm$0.000 & \textbf{0.944$\pm$0.000} & 0.885$\pm$0.000 \\
 & PSS & 0.862$\pm$0.000 & 0.887$\pm$0.000 & 0.869$\pm$0.000 & 0.879$\pm$0.000 & 0.904$\pm$0.001 & 0.912$\pm$0.000 & \textbf{0.942$\pm$0.000} & 0.929$\pm$0.000 \\
 & POS & 0.876$\pm$0.000 & 0.908$\pm$0.000 & 0.889$\pm$0.000 & 0.906$\pm$0.001 & 0.887$\pm$0.000 & 0.903$\pm$0.000 & \textbf{0.959$\pm$0.000} & 0.911$\pm$0.000 \\
\bottomrule
\end{tabular}
}
\vspace{-4mm}
\end{table*}

\begin{figure*}[t]
    \centering
    \includegraphics[width=\textwidth]{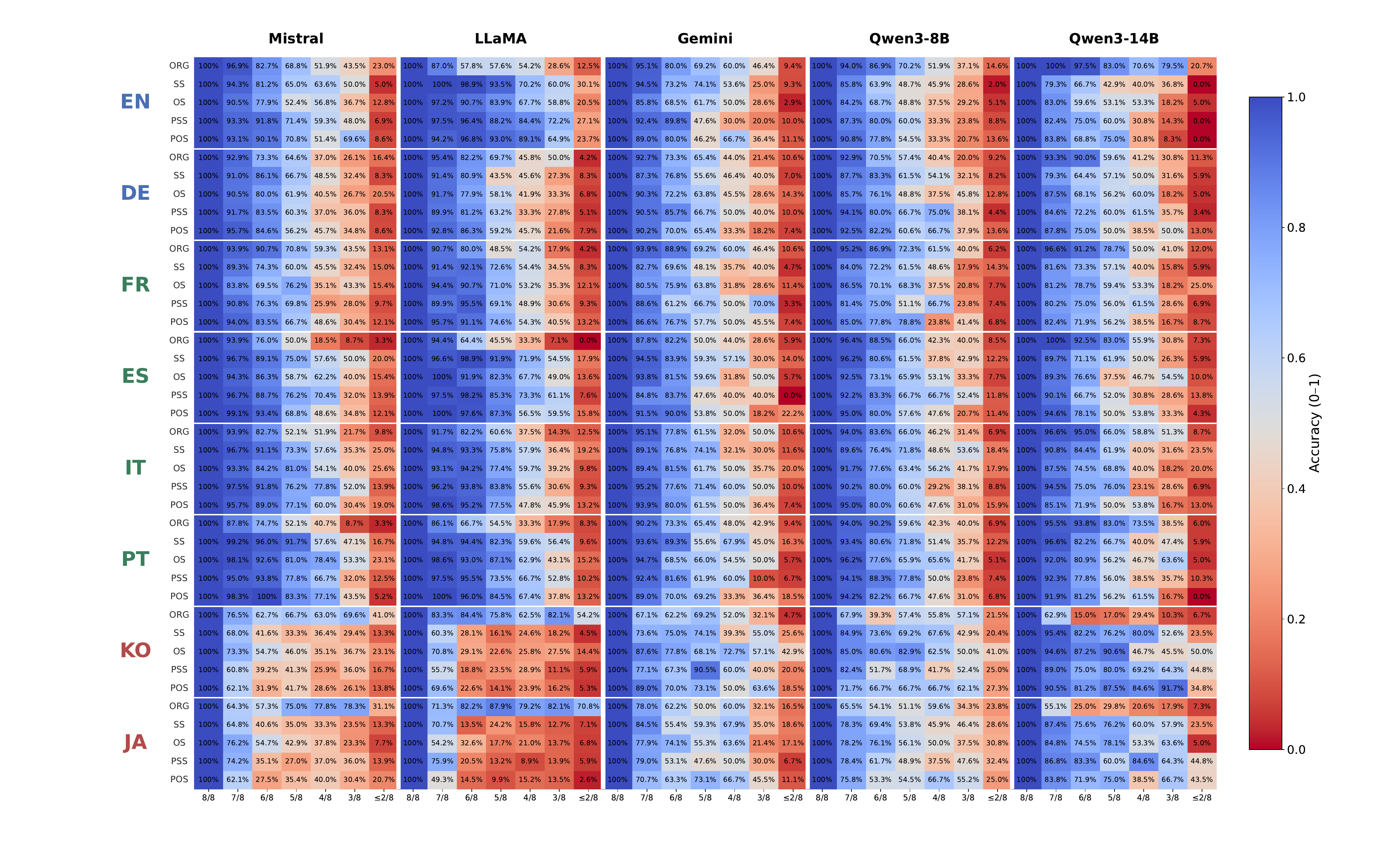}
    \caption{Accuracy distribution across models and languages grouped by problem difficulty, constructed in the same manner as Figure~\ref{fig:heatmap}, except that Qwen3 models are evaluated in think-mode.}
    \label{fig:heatmap_think}
\end{figure*}

\begin{figure*}[t]
    \centering
    \includegraphics[width=\textwidth]{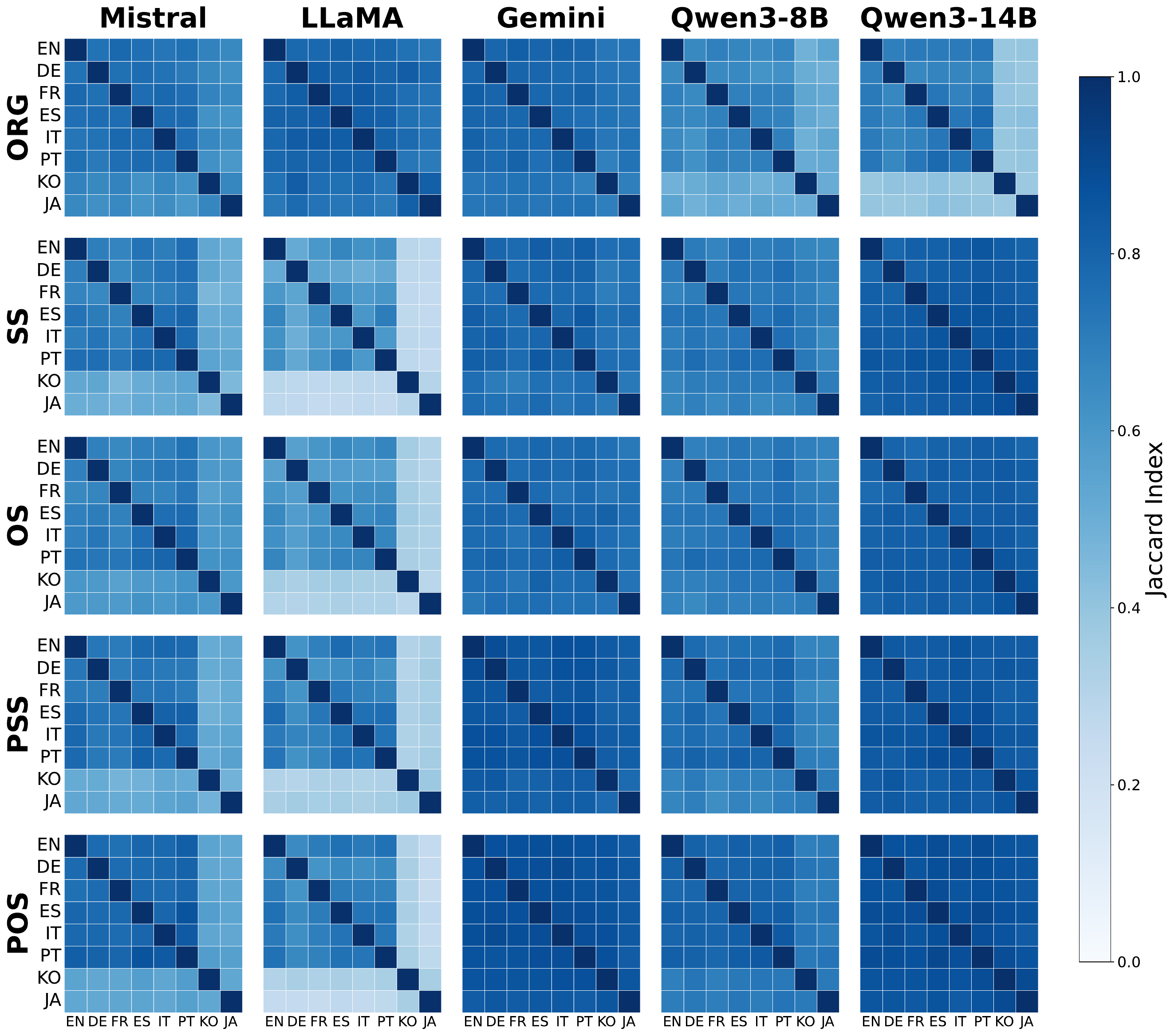}
    \caption{Language-language Jaccard similarity, constructed as in Figure~\ref{fig:jaccard_lang}, except that Qwen3 models are evaluated in think-mode. The same interpretation holds: LLaMA 3.1 and Mistral Nemo develop stronger Western/Asian language-family blocks under distortions, while Gemini and Qwen3 remain comparatively stable.}
    \label{fig:jaccard_lang_think}
\end{figure*}

\begin{figure*}[t]
    \centering
    \includegraphics[width=\textwidth]{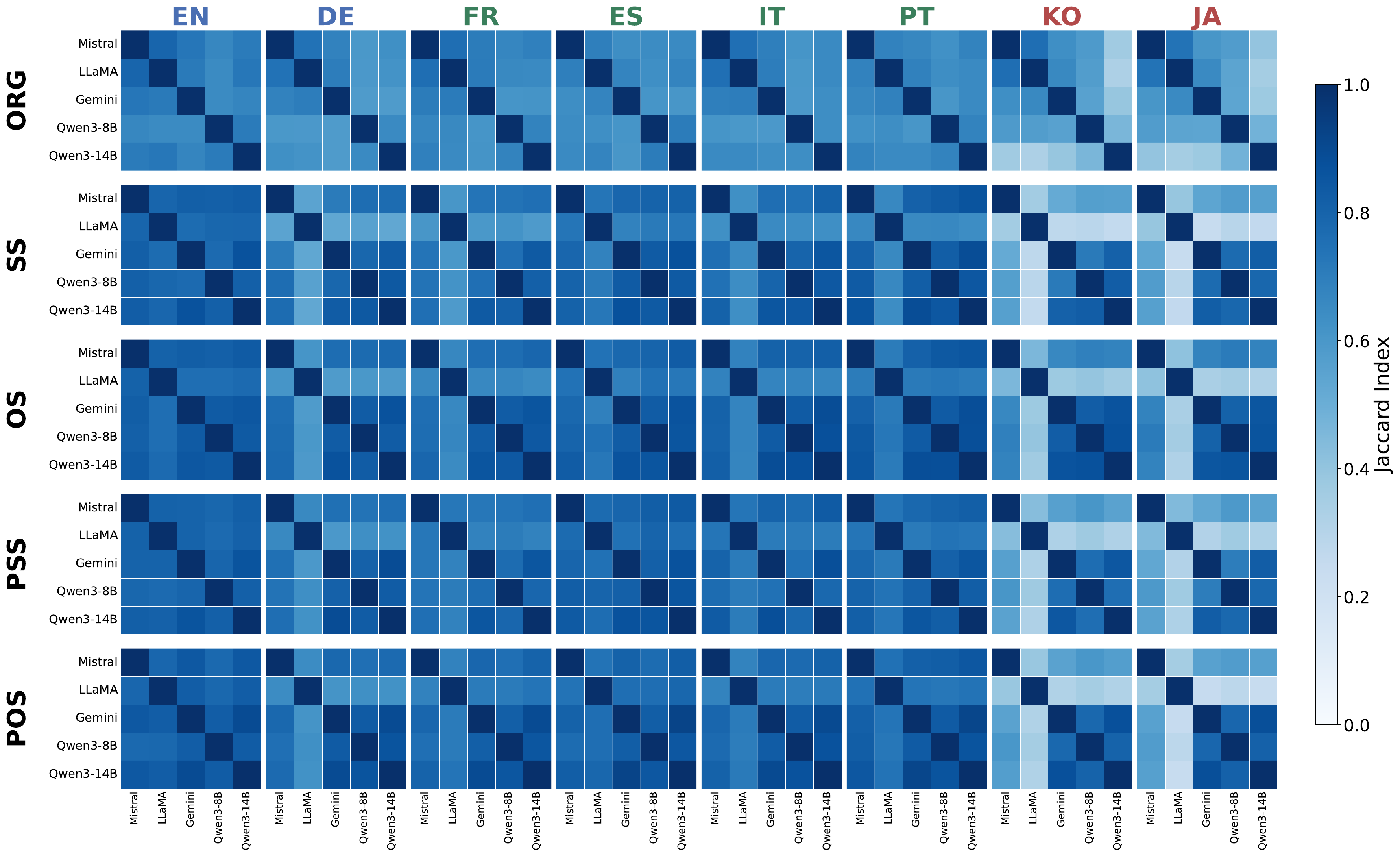}
    \caption{Model-model Jaccard similarity, constructed as in Figure~\ref{fig:jaccard_model}, except that Qwen3 models are evaluated in think-mode. The same interpretation holds: under distortions, LLaMA 3.1 shows lower agreement with other models, most prominently in Korean and Japanese.}
    \label{fig:jaccard_model_think}
\end{figure*}

\begin{table*}[t]
\centering
\caption{Arabic-only accuracy on the Arabic-label-available subset. Arabic-missing instances are removed using the union of missing row IDs across all benchmark sets. Gemini was evaluated once using the default Gemini API configuration, where dynamic thinking is enabled by default.}
\label{tab:arabic_only_results}
\vspace{-2mm}
\scriptsize
\setlength{\tabcolsep}{3pt}
\renewcommand{\arraystretch}{0.82}

{\small\textbf{(a) Unsure responses are counted as incorrect.}}

\vspace{0.5mm}

\resizebox{1.5\columnwidth}{!}{%
\begin{tabular}{llccccc}
\toprule
\textbf{Model} & \textbf{Setting} & \textbf{ORG} & \textbf{SS} & \textbf{OS} & \textbf{PSS} & \textbf{POS} \\
\midrule
Mistral-Nemo & Non-think & \textbf{0.685} & 0.630 & \textbf{0.685} & 0.603 & 0.636 \\
LLaMA-3.1-8B & Non-think & \textbf{0.864} & 0.304 & 0.217 & 0.321 & 0.266 \\
Gemini-2.5-flash & Default & 0.652 & 0.772 & 0.734 & \textbf{0.815} & \textbf{0.815} \\
Qwen3-8B & Non-think & 0.364 & 0.875 & 0.870 & \textbf{0.891} & 0.875 \\
Qwen3-8B & Think & 0.364 & 0.875 & 0.870 & \textbf{0.891} & 0.875 \\
Qwen3-14B & Non-think & 0.234 & 0.641 & 0.647 & \textbf{0.679} & \textbf{0.679} \\
Qwen3-14B & Think & 0.234 & 0.641 & 0.647 & \textbf{0.679} & \textbf{0.679} \\
\bottomrule
\end{tabular}
}

\vspace{1.5mm}

{\small\textbf{(b) Unsure responses are excluded from scoring.}}
\vspace{0.5mm}

\resizebox{1.5\columnwidth}{!}{%
\begin{tabular}{llccccc}
\toprule
\textbf{Model} & \textbf{Setting} & \textbf{ORG} & \textbf{SS} & \textbf{OS} & \textbf{PSS} & \textbf{POS} \\
\midrule
Mistral-Nemo & Non-think & \textbf{0.685} & 0.630 & \textbf{0.685} & 0.603 & 0.636 \\
LLaMA-3.1-8B & Non-think & \textbf{0.864} & 0.304 & 0.217 & 0.321 & 0.266 \\
Gemini-2.5-flash & Default & 0.678 & 0.840 & 0.828 & \textbf{0.857} & 0.852 \\
Qwen3-8B & Non-think & 0.366 & 0.875 & 0.870 & \textbf{0.891} & 0.875 \\
Qwen3-8B & Think & 0.366 & 0.875 & 0.870 & \textbf{0.891} & 0.875 \\
Qwen3-14B & Non-think & 0.331 & 0.915 & 0.915 & \textbf{0.969} & 0.912 \\
Qwen3-14B & Think & 0.331 & 0.915 & 0.915 & \textbf{0.969} & 0.912 \\
\bottomrule
\end{tabular}
}

\vspace{-3mm}
\end{table*}

\end{document}